\documentclass[11pt]{article}

\usepackage[final]{acl}

\usepackage{times}
\usepackage{latexsym}

\usepackage[T1]{fontenc}

\usepackage[utf8]{inputenc}

\usepackage{microtype}

\usepackage{inconsolata}

\usepackage{graphicx}

\usepackage{amssymb}
\usepackage{algorithm}
\usepackage{algorithmic}
\usepackage{float}
\usepackage{amsmath}
\usepackage{natbib}
\usepackage{multirow}
\usepackage{array}
\usepackage{adjustbox}
\usepackage{xcolor}
\usepackage{booktabs}

\newcommand{\name}{FAMPWQ}
\title{\name{}: Fisher Information-based Adaptive Mixed Precision Weight Quantization for Effective LLM Inference}

\author{
  \setcounter{footnote}{1}
  Gongwei Lee$^{1,3}$\thanks{\ Equal contribution. },
  \setcounter{footnote}{0}
  Ji Liu$^{2,3\dag}$\thanks{\ Corresponding author: Ji Liu (jiliuwork@gmail.com)},
  Juncheng Jia$^{1}$,
  Ji Wu$^{2}$\\
  \textsuperscript{1}School of Computer Science and Technology, Soochow University, Suzhou, China \\
  \textsuperscript{2}Electronic Engineering, Tsinghua University, Beijing, China \\
  \textsuperscript{3}Hithink Research, Hangzhou, China \\
}

\begin{document}
\maketitle
\begingroup
\endgroup
\begin{abstract}
Recent years have witnessed remarkable achievements of Large Language Models (LLMs) in multiple domains, while the excessive resource requirements of LLMs hinder the deployment on resource-constrained devices. Although model quantization stands out as an effective approach, conventional quantization approaches typically incur severe performance degradation due to uniform bit-width or simple heuristic sensitivity evaluation. In this paper, we propose a novel Fisher information-based Adaptive Mixed Precision Weight Quantization approach, i.e., \name{}, which performs layer-adaptive weight quantization for effective LLM inference on commodity GPUs. First, we propose a system model with a novel Fisher information metric to measure the layer-wise sensitivity to quantization. Second, we propose a reinforcement learning-based bit-width allocator in \name{}, which generates an adaptive bit-width allocation strategy based on the Fisher information sensitivity metric. Extensive experiments on 7 models and 5 benchmarks demonstrate that \name{} significantly outperforms 7 baseline approaches in terms of PPL (up to 3.39 smaller), accuracy (up to 6.87\% higher), and LLM-as-a-judge comparison (up to 76\% win rate).
\end{abstract}

\section{Introduction}

Recent years have witnessed remarkable progress of Large Language Models (LLMs) across a wide range of applications \cite{radford_2019_language,Petroni_2019_Languageb,Brown_2020_Languagea}. Most state-of-the-art LLMs are built upon the Transformer architecture \cite{vaswani2017attention}, achieving strong performance by scaling model size to hundreds billions \cite{agarwal2025gpt} or trillions \cite{xu2026deepseekV4} of parameters.

However, the prohibitive resource requirements of LLMs hinder their deployment on resource-constrained devices, such as edge devices, consumer GPUs, or even inference GPUs. For instance, LLaMA \cite{touvron_2023_llama,touvron2023llama,dubey2024llama} spans from 7B to 405B, which may consume from 26GB to 1500GB memory with FP32, and up to 750GB memory with FP16. Similarly, the scale of Qwen \cite{bai_2023_qwen,team2024qwen2,yang2025qwen3} can reach up to 235B corresponding to 435GB memory with FP16. In addition, DeepSeek-V4 \cite{xu2026deepseekV4} goes even larger: its 1.6T-parameter MoE (49B activated) requires about 865GB in mixed FP4/FP8, which significantly exceeds the memory of GPUs. In addition, the memory requirement scales from linear to quadratic with the sequence length. To deploy LLMs on resource-constrained devices, model quantization stands out as an effective approach.

\begin{figure}[t]
\centering
\includegraphics[width=\linewidth]{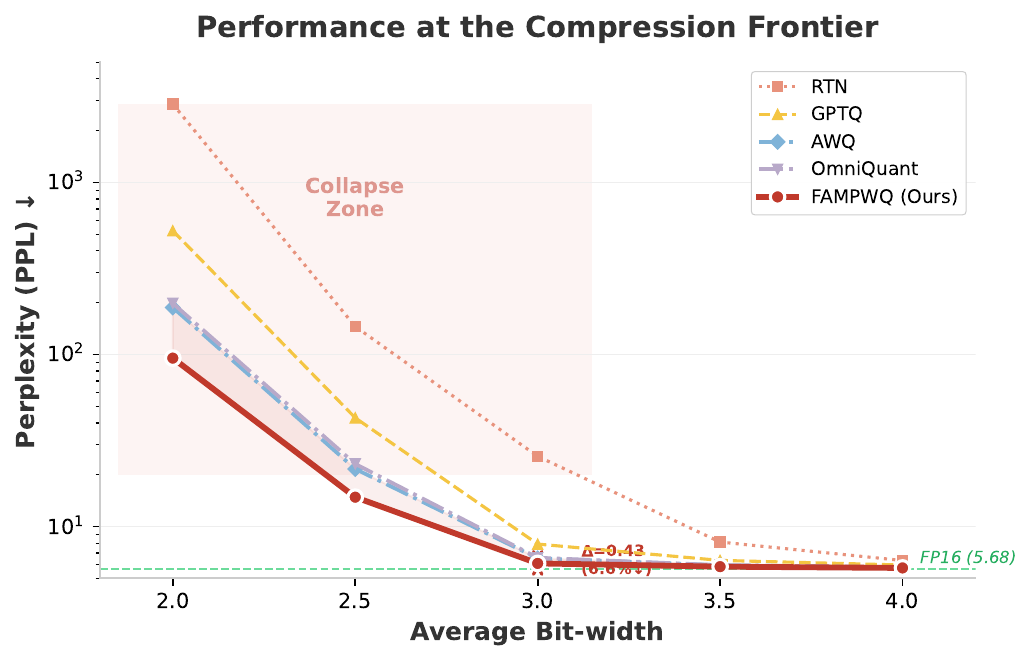}
\captionsetup{width=\linewidth}
\vspace{-8mm}
\caption{Compression-frontier behavior on LLaMA-7B with WikiText-2. \name{} keeps PPL lower as the average bit-width approaches 3 bits, where uniform quantization methods degrade rapidly.}
\label{fig:collapse_frontier}
\vspace{-4mm}
\end{figure}

Post-Training Quantization (PTQ) approaches \cite{zhu2024survey} directly quantize pre-trained LLMs without architectural modifications or retraining, albeit typically incurring performance degradation. However, existing PTQ approaches generally recognize the heterogeneous importance distribution of model weights \cite{Dong_2019_HAWQa,Gong_2024_Whatb,Wei_2022_Outlier}, with their key differentiation stemming from the statistical approaches exploited to identify and preserve critical weights. While some existing quantization approaches, e.g., GPTQ \cite{Frantar_2023_GPTQb} and AWQ \cite{Lin_2024_AWQ}, successfully reduce memory consumption through fixed bit-widths and outlier optimization, they nevertheless suffer from two fundamental limitations. First, their uniform bit-width allocation overlooks crucial layer-wise sensitivity variations, particularly in attention layers. Second, their localized outlier handling fails to account for global importance patterns across the LLM. As a consequence, the existing PTQ approaches may bring unacceptably severe performance degradation in real-life scenarios. 

While some mixed-precision approaches, e.g., OWQ \cite{Lee_2024_OWQb} and AMQ \cite{lee2025amq}, attempt to address layer heterogeneity, they rely on coarse heuristics such as weight magnitude or raw gradient norms that fail to faithfully reflect quantization-induced degradation. 

A fundamental challenge lies in the \emph{heterogeneous sensitivity} of LLM layers to quantization. Empirically, we find that certain layers (particularly \textbf{attention value projections} and \textbf{MLP down-projections}) are orders of magnitude more sensitive than others. This reveals even a small number of aggressively quantized sensitive layers can disproportionately degrade model quality, while many redundant layers can tolerate extreme compression with negligible impact. Accurately identifying which layers are critical therefore becomes the key to effective mixed-precision quantization.

Existing sensitivity metrics~\cite{Frantar_2022_Optimal,Lin_2024_AWQ}, however, are ill-suited to this task. Weight magnitude and gradient norms capture only first-order statistics and do not reflect the geometry of the loss surface under quantization-specific perturbations. Second-order point estimates, including Hessian-based~\cite{Dong_2019_HAWQa} and standard Fisher-based metrics, evaluate curvature only at the unperturbed weights and remain agnostic to the bit-width-specific noise that quantization actually injects. To bridge this gap, we propose a \emph{perturbation-based Fisher Information} metric that directly injects quantization-simulating perturbations into layer weights and measures the resulting shift in the FIM. Different from these point-estimate metrics, our formulation captures how quantization noise, rather than arbitrary parameter variations, distorts the local loss geometry, providing a principled and quantization-specific layer sensitivity measure.

\begin{figure}[!t]
\centering
\includegraphics[width=\linewidth]{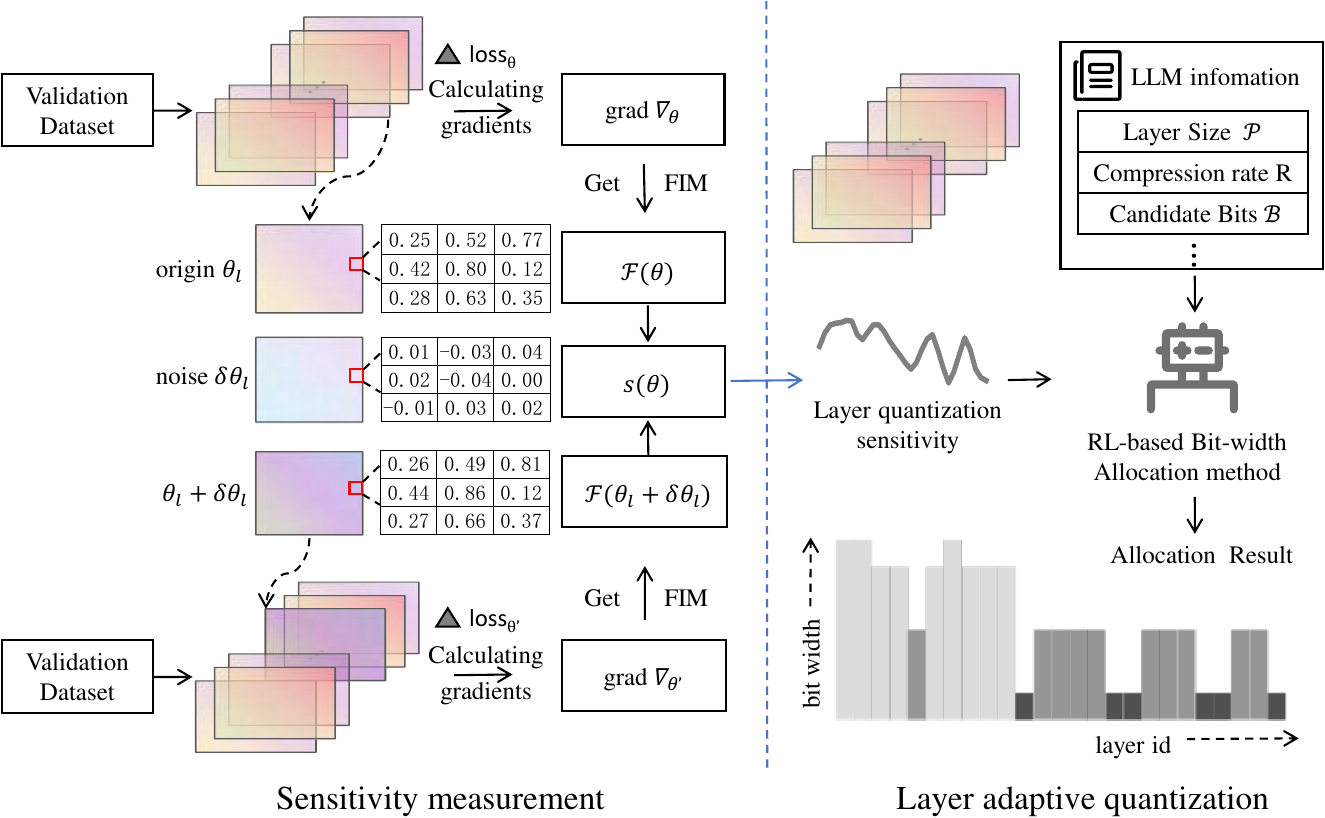}
\captionsetup{width=\linewidth}
\vspace{-6mm}
\caption{System model of \name{}.}
\label{fig:framework}
\vspace{-4mm}
\end{figure}

In this paper, we propose a Fisher Information-based Adaptive Mixed Precision Weight Quantization (\name{}) approach, i.e., a weight-only mixed-precision PTQ framework for fixed-memory LLM deployment. \name{} introduces a quantization-perturbation Fisher sensitivity metric that estimates per-layer degradation than magnitude/gradient proxies, and exploits a low-cost proxy optimizer to allocate layer bitwidths under a storage budget. As shown in Figure \ref{fig:framework}, \name{} consists of two stages: a perturbation-based Fisher sensitivity measurement stage and a Reinforcement Learning (RL)-based adaptive bit-width allocation stage. As shown in Figure~\ref{fig:collapse_frontier}, while uniform PTQ approaches are competitive around 4 bits,  their PPL rises sharply below 3.5 average bits. By preserving sensitive layers and compressing tolerant layers more aggressively, \name{} maintains significantly lower degradation below the 3-bit frontier.

The major contributions are as follows:
\begin{enumerate}
    \item We propose a system model with a novel sensitivity measurement method based on a new Fisher Information metric for layer adaptive quantization. The Fisher Information metric explicitly injects quantization-simulating perturbations and measures the resulting Fisher shift to capture layer-wise sensitivity to quantization loss. 
    \item We propose an adaptive bit-width allocator in \name{} to enable storage-constrained mixed-precision search guided by quantization-specific loss geometry. The allocator generates an adaptive bit-width allocation strategy based on Proximal Policy Optimization (PPO) and the quantization perturbation Fisher sensitivity of each layer, for layer-wise quantization of LLMs.
    \item We implement \name{} and maintain the compatibility with existing methods, e.g., GPTQ or AWQ. We carry out extensive experiments on 7 models and 5 benchmarks to demonstrate that \name{} significantly outperforms 7 baseline approaches in terms of PPL (up to 3.39 smaller), accuracy (up to 6.87\% higher), and LLM-as-a-judge comparison (up to 76\% win rate).
\end{enumerate}

\section{Related Work}

Recent LLM post-training quantization (PTQ) studies improve compression by reducing quantization error, correcting outliers, or calibrating quantized weights. SmoothQuant \cite{Xiao_2023_SmoothQuantb} redistributes quantization difficulty between weights and activations, GPTAQ \cite{Li_2025_GPTAQa} mitigates error accumulation through calibration, and OmniQuant \cite{Shao_2024_OmniQuant} and ABQ-LLM \cite{Zeng_2025_ABQ-LLM} further explore adaptive clipping and bit-balance strategies for low-bit settings. Other methods, such as SqueezeLLM \cite{Kim_2024_SqueezeLLMa} and OWQ \cite{Lee_2024_OWQb}, preserve salient weights or channels at higher precision. These works show the importance of protecting sensitive parameters, but their adaptation is generally local and does not directly optimize layer-wise precision under a global memory budget.

This limitation has motivated mixed-precision and search-based allocation. DeepSeek-V4 models \cite{xu2026deepseekV4} adopt mixed FP4/FP8 expert-aware quantization for MoE deployment at expert granularity. AMQ \cite{lee2025amq} exploits activation-guided mixed precision, HAQ \cite{wang2019haq} employs reinforcement learning for hardware-aware bit-width search, COPAL \cite{malla2024copal} formulates layer-wise allocation as combinatorial optimization, and BitWeaver \cite{gagnon2025bitweaver} investigates hardware-efficient mixed-precision layouts. RL-PTQ \cite{wang2024rl} applies reinforcement learning, but depends on repeated model-level evaluation. \name{} focuses on layer-wise sensitivity differences under a global memory budget; it estimates layer sensitivity via Fisher Information and employs a proxy-guided PPO allocator to efficiently search storage-constrained bit-width configurations, achieving superb performance.

\section{System Model and Problem Formulation}

In this section, we present the system model of \name{}, and formulate the problem to address in LLM quantization.

\subsection{System Model}

\begin{figure}[!t]
\centering
\includegraphics[width=\linewidth]{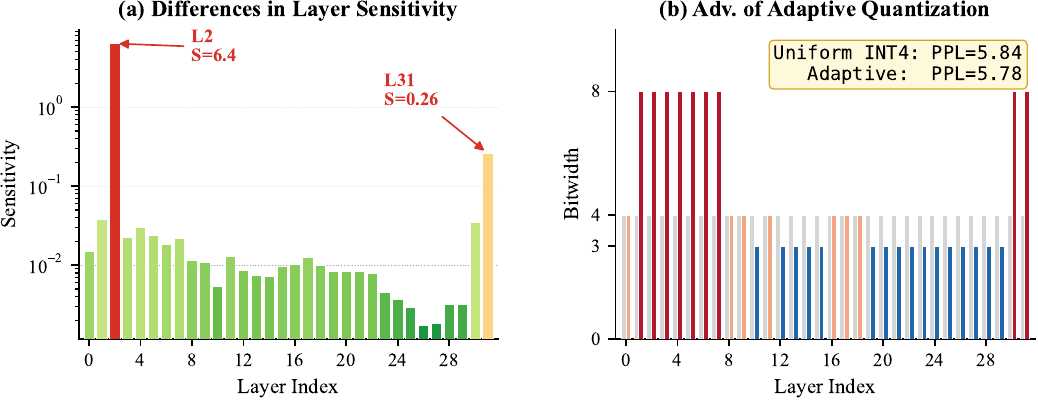}
\captionsetup{width=\linewidth}
\vspace{-8mm}
\caption{Motivation for adaptive bitwidth quantization.}
\label{fig:motivation}
\vspace{-6mm}
\end{figure}

While conventional quantization approaches employ identical bit-widths for all layers, LLM layers exhibit highly uneven tolerance to quantization. As shown in Figure~\ref{fig:motivation}(a), the measured layer sensitivity spans orders of magnitude within the same model, indicating that a few critical layers can dominate quantization-induced degradation. Figure~\ref{fig:motivation}(b) further shows that, under a comparable average precision budget, a mixed-precision allocation achieves lower WikiText-2 PPL than uniform INT4 by assigning higher bit-widths to sensitive layers and lower bit-widths to tolerant ones. These observations motivate a layer-adaptive quantization framework that explicitly measures sensitivity and allocates precision under a global storage constraint.

As shown in Figure~\ref{fig:framework}, the system model of \name{} consists of two stages: sensitivity measurement and layer-adaptive quantization. In the sensitivity measurement stage, we inject perturbation noise into layer weights and compute the Fisher information change to quantify each layer's sensitivity (Section \ref{subse:fisherinfo}). In the layer-adaptive quantization stage, an RL-based method allocates appropriate bit-widths to each layer based on the computed sensitivity (Section \ref{subsec:bitallo}). The resulting allocation strategy is then applied via any compatible quantization method (e.g., AWQ).




\subsection{Problem Formulation}

Let us consider an LLM $M$ composed of $L$ layers. The sensitivity value of Layer $l$ is denoted by $s_l, l \in [1, L]$. In order to quantize the LLM, we define a set of available bit-width options, $\mathcal{B}$, comprising one or more discrete choices (e.g., $\{4,16\}$ or $\{2,3,4,8\}$). We assume that the sensitivity of each layer is independent of its allocated bit-width $q_l$. Then, the accuracy degradation incurred by quantizing an individual layer $\Delta Acc_l$ is proportional to its sensitivity and decreases exponentially with increasing bit-width \cite{zhou2018adaptive}, as shown in Formula \ref{eq:accuracy_loss2single_layer}:
\begin{equation}\label{eq:accuracy_loss2single_layer}
  \begin{aligned}
    \Delta Acc_l(\mathcal{Q}) \propto s_l \cdot e^{-\alpha (q_l/B)},
  \end{aligned}
\end{equation}
where $B$ represents the original (full-precision) bit-width, $\mathcal{Q} = \{(1, q_1), ..., (L, q_L)\}$ refers to a quantization bit-width allocation strategy, $\alpha$ is a positive constant decay rate, and $e$ is Euler's number. We define the accuracy degradation by normalizing the exponential decay fluctuation induced by bit-width $q_l$ as Formula \ref{eq:accuracy_loss2single_layerb}.
\begin{equation}\label{eq:accuracy_loss2single_layerb}
  \begin{aligned}
    \Delta Acc_l(\mathcal{Q}) = \frac{Acc_o \cdot s_l}{\sum_{i=1}^L s_i} \cdot \frac{(e^{-\alpha (q_l/B)}-e^{-\alpha})}{(1-e^{-\alpha})},
  \end{aligned}
\end{equation}
where $Acc_o$ represents the original accuracy of LLM $M$ without quantization. Afterwards, we can calculate the total accuracy degradation brought by all layers as defined in Formula \ref{eq:accuracy_loss2total_model}.
\begin{equation}\label{eq:accuracy_loss2total_model}
  \begin{aligned}
    \Delta Acc(\mathcal{Q}) = \frac{Acc_o}{\sum_{i=1}^L s_i} \sum_{l=1}^{L}s_l \cdot \frac{(e^{-\alpha (q_l/B)}-e^{-\alpha})}{(1-e^{-\alpha})}.
  \end{aligned}  
\end{equation}

The problem we address in this work is how to find a bit-width allocation strategy $\mathcal{Q}^*$ so as to minimize the accuracy degradation while achieving the compression rate target as formulated in Formula \ref{eq:objective}.
\begin{equation}\label{eq:objective}
  \begin{aligned}
    &\mathcal{Q}^* = \operatorname*{argmin}_{\mathcal{Q}} \Delta Acc(\mathcal{Q}), \\
    \text{s.t.} &\begin{cases} 
    \forall \quad (l, q_l) \in \mathcal{Q}^*, q_l \in \mathcal{B},\\
    \sum_{l=1}^L p_l q_l \leq R \cdot B \cdot \sum_{l=1}^L p_l,
    \end{cases} 
  \end{aligned}  
\end{equation}
where $p_l$ is the number of parameters in Layer $l$, and $R$ is the target compression ratio. This problem definition bridges the accuracy and memory requirement by optimizing bit-width allocation strategy $\mathcal{Q}$, where the objective function $\Delta Acc(\mathcal{Q})$ explicitly represents the accuracy degradation, while the compression rate target guarantees hardware compatibility in terms of memory requirement. This problem is complicated due to severe combinatorial explosion. The search space grows exponentially as $\mathcal{O}(|\mathcal{B}|^L)$. For instance, the search space reaches $3^{224} \approx 10^{106}$ for LLaMA3-8B (224 layers from 32 blocks × 7 layers) with only 3 bit-width options for each layer, rendering exhaustive search computationally prohibitive even for offline quantization.


    

\section{\name{} Methodology}

In this section, we detail the methodology of \name{}. We first describe the Fisher information-based sensitivity measurement for each layer. Then, we present the adaptive bit-width allocation method that minimizes accuracy degradation while achieving the compression rate target.

\subsection{Fisher Information-based Sensitivity}
\label{subse:fisherinfo}

Fisher information quantifies the amount of information that observable data carries about unknown model parameters. We leverage this property to measure the sensitivity of each layer to quantization noise: a layer whose Fisher information changes substantially under perturbation is highly sensitive. Specifically, we compute the Fisher information of each layer with its original weights and with perturbed weights, and use the difference as the sensitivity measure.

    
    
In order to quantify layer sensitivity, we inject a perturbation $\delta\theta_l$ into the parameters of each Layer $l$ and measure the resulting shift in the FIM. While a generic perturbation, e.g., uniform or magnitude-proportional noise, only reflects general parameter importance, we need to capture the \emph{specific} noise incurred by $b$-bit quantization. We therefore exploit Formula \ref{eq:delta3} to generate the perturbation.
\begin{equation}\label{eq:delta3}
\delta\theta_l = Q_b(\theta_l) - \theta_l,
\end{equation}
where $Q_b(\cdot)$ denotes the $b$-bit quantize, i.e., the dequantization operator. This definition ensures $\theta_l + \delta\theta_l = Q_b(\theta_l)$, so $\delta\theta_l$ is exactly the additive rounding perturbation introduced by $b$-bit quantization rather than an arbitrary direction.

To simplify sensitivity evaluation, we adopt a layer-independent strategy. We add the perturbation to only one target layer $\theta_l$, while all other layers remain at their original parameters. Then, we can get the layer after adding perturbation noise as defined in Formula \ref{eq:addNoise}.
\begin{equation}\label{eq:addNoise}
\theta_l^{\text{pert}} = \theta_l + \delta\theta_l,
\end{equation}
where $\theta_l$ refers to the original parameters of Layer $l$ and $\theta_l^{\text{pert}}$ is the parameters with perturbation.

\begin{figure}[!t]
\centering
\includegraphics[width=\linewidth]{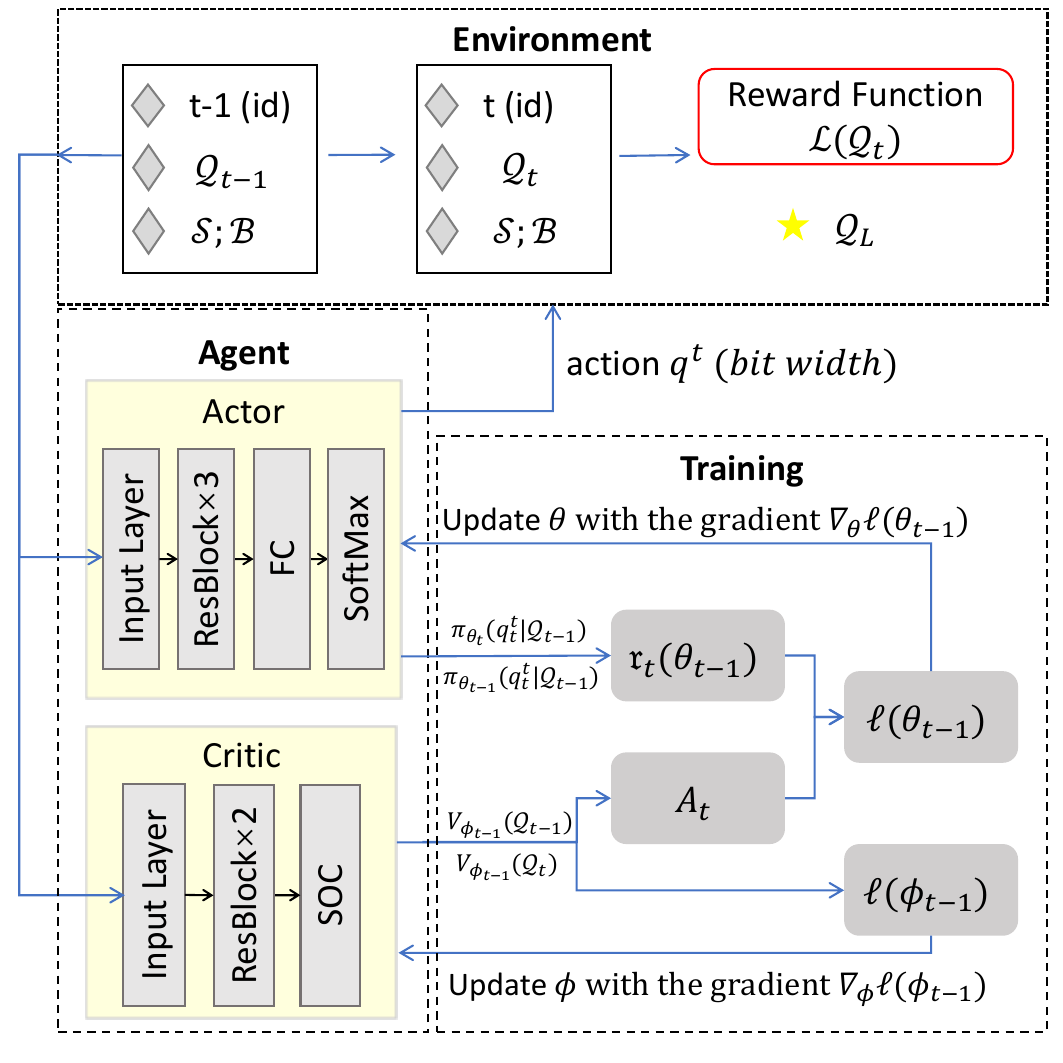}
\captionsetup{width=\linewidth}
\vspace{-8mm}
\caption{PPO-based adaptive bit-width allocation method. $\ell(\theta_{t-1}) = \mathbb{E} \left[ \min\left( \mathfrak{r}_t(\theta_{t-1})A_t, \mathfrak{c}_t(\theta_{t-1})A_t \right) \right]$ and $\ell(\phi) = A_t^2$, which are exploited in Formulas \ref{eq:upgrade_theta} and \ref{eqn:upgrade_phi}. $\mathcal{S} = \{s_1, ..., s_L \}$ represents the sensitivity.}
\label{fig:rlArch}
\vspace{-6mm}
\end{figure}

We can calculate the gradients $\nabla_{\theta_l} \ell(x|\theta)$, which denotes the first-order derivative of the LLM. Then, we can derive the empirical Fisher Information Matrix (FIM) as defined in Formula \ref{eq:fisher_emirical}. 
\begin{equation}\label{eq:fisher_emirical}
    \hat{F}(\theta_l) = \frac{1}{N}\sum^N_{n=1} (\nabla_{\theta_l}\ell(x_n|\theta) \nabla_{\theta_l}\ell(x_n|\theta)^\top).
\end{equation}
We use the diagonal vector of FIM denoted by $\mathcal{F}(\theta_l)$ to approximate the original FIM \cite{Liu_2024_Fishera} as defined in Formula \ref{eq:FIMErank}.
\begin{equation}\label{eq:FIMErank}
    \mathcal{F}(\theta_l) = \text{diag}(I_{|\nabla_{\theta_l}\ell(x|\theta)|}\odot \hat{F}(\theta_l)).
\end{equation} 
Since the diagonal vector of FIM only depends on the diagonal elements of the original matrix, we can simplify the calculation of Formulas \ref{eq:fisher_emirical} and \ref{eq:FIMErank}. We can calculate each element $f_i(\theta_l) \in \mathcal{F}(\theta_l)$ with $i$ representing the $i$-th element in $\mathcal{F}(\theta_l)$ as defined in Formula \ref{eq:simpleFIM}.
\begin{equation}\label{eq:simpleFIM}
    f_i(\theta_l) = \frac{1}{N}\sum^N_{n=1} \sum^J_{j=1} (\nabla_{\theta_l}\ell(x_n|\theta))_{(i,j)}^2,
\end{equation} 
where $\nabla_{\theta_l}\ell(x_n|\theta)_{(i,j)}$ represents the element with the index ($i$, $j$) in $\nabla_{\theta_l}\ell(x_n|\theta)$ and $J$ represents the number of elements in $i$-th row of $\nabla_{\theta_l}\ell(x_n|\theta)$. Similarly, we can calculate the diagonal vector of FIM for the parameters with added perturbation noise denoted by $\mathcal{F}(\theta_l^{\text{pert}})$. See calculation details of FIM in Appendix \ref{AppSubSubSec:FIM}.

Finally, we take the FIM variation to indicate the sensitivity of a layer, which is defined in Formula \ref{eq:sensitivity}.
\begin{equation}\label{eq:sensitivity}
    {s}_{l} =
    \frac{\left\|\mathcal{F}(\theta_l^{\text{pert}}) -
    \mathcal{F}(\theta_l)\right\|_2}
    {\left\|\mathcal{F}(\theta_l)\right\|_2},
\end{equation} 
where ${s}_{l}$ represents the sensitivity of Layer $l$, $||\cdot||_2$ is the Euclidean norm over the diagonal FIM vector. The resulting scalar sensitivity value $s_l$ is used in Formula \ref{eq:accuracy_loss2total_model}.

\subsection{Adaptive Bit-width Allocation}
\label{subsec:bitallo}

In this section, we present an RL-based adaptive bit-width allocation method. Since the combinatorial problem defined in Formula \ref{eq:objective} is intractable, we transform it into a single loss function minimization problem as defined in Formula \ref{eq:rl_objective}.
\begin{equation}\label{eq:rl_objective}
    \min \mathcal{L} (\mathcal{Q}) =  \Delta Acc(\mathcal{Q}) + P(\psi (\mathcal{Q})) \cdot || \psi (\mathcal{Q}) ||^2,
\end{equation}
where $P(\psi (\mathcal{Q}))$ is a penalty parameter and $\psi(\mathcal{Q})$ is the storage loss compared with the compression rate target $R$ as defined in Formula \ref{eq:compression}.
\begin{equation}\label{eq:compression}
    \begin{aligned}
    \psi(\mathcal{Q}) = \sum_{i=1}^{L}p_i\cdot q_i - R \cdot B \cdot \sum_{i=1}^L p_i.
    \end{aligned}
\end{equation}
In addition, $P(\psi (\mathcal{Q}))$ depends on $\psi(\mathcal{Q})$ as defined in Formula \ref{eq:penalty}.
\begin{equation}\label{eq:penalty}
    \begin{aligned}
    P(\psi (\mathcal{Q})) = \begin{cases} 
    P_{\text{penalty}} > 0, \text{if }\psi(\mathcal{Q}) > 0,\\
    P_{\text{reward}} \le 0, \text{otherwise},
    \end{cases} 
    \end{aligned}
\end{equation}
where $P_{\text{penalty}}$ is the penalty when the quantization does not achieve the targeted compression rate and $P_{\text{reward}}$ refers to the rewards brought by the extra quantization compression. Both $P_{\text{penalty}}$ and $P_{\text{reward}}$ are constant values.

While RL is an effective approach for complex combinatorial optimization problems \cite{cappart2021combining}, we adopt Proximal Policy Optimization (PPO) \cite{schulman_2017_proximal} for bit-width allocation. As shown in Figure \ref{fig:rlArch}, the architecture consists of an agent and the environment. The agent generates the bit-width allocation strategy while the environment provides feedback through a reward function. The agent consists of two modules: the actor generates bit-width allocation strategies and the critic guides policy optimization. Both the actor and critic modules are implemented as lightweight residual networks (see architecture details in Appendix \ref{AppSubSec:rlTraining}). During the quantization phase, both modules are first trained, after which the actor generates the final allocation strategy. The actor takes the layer ID, the current allocation $\mathcal{Q}$, per-layer parameter counts, per-layer sensitivity scores, and the candidate bit-widths $\mathcal{B}$ as input. It then outputs the bit-width for the corresponding layer. The critic receives the same inputs and produces a scalar value estimate to guide policy optimization.

\subsubsection{Training Process}

The training process contains multiple epochs, each of which consists of $L$ steps. At the beginning of the training, the bit-width allocation strategy $\mathcal{Q}$ is initialized to the highest selectable bit-width in each layer, i.e., $\forall (l, q_l) \in \mathcal{Q}_0, q_l \in \mathcal{B}$, which is exploited for the first epoch. For each epoch, at Step $t$, we denote the parameters of the actor network by $\theta_t$ and that of the critic network by $\phi_t$. We denote the bit-width for Layer $l$ at Step $t$ by $q_l^{t}$. Then, the actor network generates the bit-width $q^t_t$ for Layer $t$, and update $\mathcal{Q}_{t-1}$ to $\mathcal{Q}_{t}$ by replacing $q_t^{t-1}$ by $q_t^t$.  In addition, we denote the scalar value of the critic network by $V_{\phi_t}(\mathcal{Q}_t)$. Then, we compute the Temporal-Difference (TD) advantage \cite{rowland2024analysis} $A_t$ at Step $t$ as defined in Formula \ref{eqn:advantage}.
\begin{equation}\label{eqn:advantage}
A_t = \mathcal{L}(\mathcal{Q}_{t-1}) + \gamma V_{\phi_{t-1}}(\mathcal{Q}_t) - V_{\phi_{t-1}}(\mathcal{Q}_{t-1}),
\end{equation}
where $\mathcal{L}(\mathcal{Q}_{t-1})$ is defined in Formula \ref{eq:rl_objective}, $\gamma \in (0,1)$ is a discount factor that controls the trade-off between immediate and future rewards. The actor network is updated by minimizing the clipped surrogate objective \cite{schulman_2017_proximal} while ensuring stable policy improvements as defined in Formula \ref{eq:upgrade_theta}.
\begin{align}\label{eq:upgrade_theta}
\theta_t \leftarrow  \theta_{t-1} - \eta_\theta \nabla_{\theta_{t-1}} \mathbb{E}\big[\min\big(\mathfrak{r}_t(\theta_{t-1}) A_t, & \notag\\
\mathfrak{c}_t(\theta_{t-1}) A_t  \big) \big] &
\end{align}
where $\mathbb{E} \left[ \cdot \right]$ corresponds to the empirical average, $\eta_\theta$ is a constant learning rate of the actor network, $\mathfrak{c}_t(\theta_{t-1})$ refers to a clip reward defined in Formula \ref{eq:clip_reward}:
\begin{align} \label{eq:clip_reward}
\mathfrak{c}_t(\theta_{t-1}) = \text{clip}(\mathfrak{r}_t(\theta_{t-1}), 1-\epsilon, 1+\epsilon),
\end{align}
where $\epsilon$ is a small constant controlling the policy update range, $\mathfrak{r}_t(\theta)$ represents the policy-dependent reward as defined in Formula \ref{eq:policy_reward}. 
\begin{align} \label{eq:policy_reward}
\mathfrak{r}_t(\theta_{t-1}) = \frac{\pi_{\theta_t}(q_t^t | \mathcal{Q}_{t-1})}{\pi_{\theta_{t-1}}(q_t^t | \mathcal{Q}_{t-1})},
\end{align}
where $\pi_{\theta_t}(q_t^t | \mathcal{Q}_{t-1})$ represents the probability to generate $q_t^t$ with the actor network $\theta_t$ and the allocation strategy $\mathcal{Q}_{t-1}$.

Simultaneously, the critic network is updated to minimize the squared TD advantage:
\begin{equation} \label{eqn:upgrade_phi}
\phi_t \leftarrow  \phi_{t-1} - \eta_{\phi}\nabla _{\phi_{t-1}}A_t^2,
\end{equation}
where $\eta_{\phi}$ is a constant learning rate of the critic network. See training details in Appendix \ref{AppSubSec:rlTraining}.

\subsubsection{Inference Process} The inference process consists of $L$ steps. Similar to the training process, the bit-width allocation strategy $\mathcal{Q}_{0}$ is initialized to the highest selectable bit-width. At each step $t$, the actor module generates a bit-width $q_t^t$ for Layer $t$ and updates $\mathcal{Q}_{t-1}$ by replacing $q_t^{t-1}$ with $q_t^t$. After $L$ steps, $\mathcal{Q}_{L}$ contains the generated bit-widths and is used as the adaptive allocation strategy to quantize the LLM.

\section{Experiments}

In this section, we present the experimental results. We first describe the experimental setup and then compare \name{} with 7 baseline approaches across 7 models and 5 benchmarks. We implement \name{} in Python while maintaining compatibility with existing quantization backends such as GPTQ, AWQ, and OmniQuant.

\begin{table}[t]
    \centering
    \vspace{-4mm}
    \caption{PPL $\downarrow$ comparison on LLaMA-7B and LLaMA-13B with 4-bit and 3-bit average quantization. \textbf{Bold} indicates the lowest PPL and \underline{underlined} indicates the second lowest.}
    \label{tab:ppl_llama}
    \vspace{-4mm}
    \renewcommand{\arraystretch}{1}
    \resizebox{\hsize}{!}{
    \begin{tabular}{l|c|cccc|cccc}
        \toprule
        \textbf{Model} &  & \multicolumn{4}{c|}{\textbf{LLaMA-7B}} &
        \multicolumn{4}{c}{\textbf{LLaMA-13B}} \\
        \midrule
        \textbf{Method} & \multicolumn{1}{c|}{Avg bit} &
        Wiki2 & PTB & C4 & Avg PPL & Wiki2 & PTB & C4 & Avg PPL\\
        \midrule
        FP16 & 16 & 5.68 & 10.11 & 7.34 & 7.71 & 5.09 & 9.08 & 6.80 & 6.99 \\
        \midrule
        RTN & 4 & 6.29 & 11.23 & 8.12 & 8.55 & 5.53 & 9.77 & 7.23 & 7.51 \\
        GPTQ & 4 & 6.01 & 10.59 & 7.74 & 8.11 & 5.30 & 9.37 & 6.96 & 7.21 \\
        GPTQv2 & 4 & 5.89 & 10.46 & 7.58 & 7.98 & 5.24 & 9.31 & 6.93 & 7.16 \\
        OmniQuant & 4 & 5.86 & \underline{10.42} & \textbf{7.53} & 7.94 &
        5.21 & 9.21 & 6.91 & 7.11 \\
        AMQ & 4 & 5.88 &  10.43 & 7.62 & 7.98 & 5.25 & 9.40 & 6.97& 7.20\\
        OWQ & 4 & 5.96 & 10.67 & 7.67 & 8.10 & 5.25 & 9.32 & 6.97 & 7.18 \\
        AWQ & 4 &
        \underline{5.83} & \underline{10.42} & \textbf{7.53} & \underline{7.93} &
        \underline{5.20} & \underline{9.20} & \underline{6.90} & \underline{7.10} \\

        \name{} & 4 &
        \textbf{5.81} & \textbf{10.34} & \underline{7.54} & \textbf{7.90} &
        \textbf{5.19} & \textbf{9.18} & \textbf{6.88} & \textbf{7.08} \\

        \midrule
        RTN & 3 & 25.58 & 89.45 & 30.81 & 48.61 & 11.40 & 26.36 & 14.38 & 17.38 \\
        GPTQ & 3 & 7.90 & 14.72 & 10.23 & 10.95 & 5.82 & 8.62 & 6.78 & 7.07 \\
        GPTQv2 & 3 & 7.31 & 12.64 & 8.97 & 9.64 & 5.68 & 8.45 & 6.70 & 6.94 \\
        OmniQuant & 3 & \underline{6.49} & \underline{11.43} & \underline{8.19} & \underline{8.70} & \underline{5.48} & \underline{8.21} & \underline{6.35} & \underline{6.68} \\
        AMQ & 3 & 6.83 & 12.66 & 8.72 & 9.40 & 5.68 & 8.51 & 6.54 & 6.91 \\
        OWQ & 3 & 6.65 & 12.47 & 8.62 & 9.25 & 5.66 & 10.02 & 7.43 & 7.70 \\
        AWQ & 3 & 6.53 & 11.83 & 8.58 & 8.98 & 5.52 & 8.31 & 6.42 & 6.75 \\
        \name{} & 3 & \textbf{6.35} & \textbf{11.33} & \textbf{8.07} & \textbf{8.58} & \textbf{5.40} & \textbf{8.20} & \textbf{6.25} & \textbf{6.62} \\

        \bottomrule
    \end{tabular}
    }
\vspace{-7mm}
\end{table}

\subsection{Experimental Setup}

We take 6 state-of-the-art PTQ approaches, i.e., GPTQ \cite{Frantar_2023_GPTQb}, GPTQv2 \cite{Li_2025_GPTAQa}, AMQ \cite{lee2025amq}, OmniQuant \cite{Shao_2024_OmniQuant}, OWQ \cite{Lee_2024_OWQb}, and AWQ \cite{Lin_2024_AWQ} as baseline approaches. We take a simple quantization approach by mapping floating-point values to their nearest discrete levels, which is denoted by Round-To-Nearest (RTN), as a baseline approach.  We evaluate on 7 LLMs, i.e., LLaMA-7B, LLaMA-13B \cite{touvron_2023_llama}, LLaMA2-7B-chat, LLaMA2-13B-chat \cite{touvron2023llama}, Qwen2.5-7B, Qwen2.5-14B \cite{qwen2_2024} and Mistral-7B-v0.1 \cite{jiang2023mistral7b}. In addition, we utilize 5 benchmarks: Wikitext-2 (Wiki2) \cite{merity_2016_pointer}, Penn Treebank (PTB) \cite{Marcus_1994_Pennb}, C4 \cite{Raffel_c4_2020}, lm-evaluation-harness \cite{eval-harness}, and Vicuna \cite{chiang_2023_vicuna}, to evaluate the PPL, the accuracy, and the LLM-as-a-judge comparison of diverse quantization approaches. 

\begin{table}[t]
    \centering
    \renewcommand{\arraystretch}{1.2}
    \caption{Zero-shot reasoning accuracy ($\uparrow$) of quantized Qwen2.5-7B under 3-bit quantization. }
    \vspace{-2mm}
    \resizebox{\hsize}{!}{
    \begin{tabular}{l|c|cccccc}
        \toprule
        \multicolumn{8}{c}{\textbf{Qwen2.5-7B}} \\
        \midrule
        \textbf{Method} & \textbf{Avg bit} & \textbf{BoolQ} & \textbf{ARC-E} & \textbf{ARC-C} & \textbf{HellaSwag} & \textbf{WinoGrande} & \textbf{Avg acc} \\
        \midrule
        FP16 (Ref) & 16 & 0.8471 & 0.8047 & 0.4778 & 0.6003 & 0.7301 & 0.6920 \\
        \midrule
        RTN & 3 & 0.6425 & 0.5851 & 0.3846 & 0.5075 & 0.5983 & 0.5436 \\
        GPTQ & 3 & 0.6845 & 0.5912 & 0.3756 & 0.4867 & 0.5891 & 0.5454 \\
        OWQ & 3 & 0.7156 & 0.6083 & 0.3821 & 0.5074 & 0.6022 & 0.5631 \\
        GPTQv2 & 3 & 0.7324 & 0.6245 & 0.3878 & 0.5192 & 0.6114 & 0.5751 \\
        OmniQuant & 3 & \underline{0.7634} & 0.6572 & \underline{0.3956} & 0.5348 & 0.6231 & 0.5948 \\
        AWQ & 3 & 0.7612 & \underline{0.6588} & 0.3941 & \underline{0.5456} & \underline{0.6245} & \underline{0.5968} \\
        \midrule
        \textbf{\name{} (Ours)} & 3 & \textbf{0.7854} & \textbf{0.6821} & \textbf{0.4032} & \textbf{0.5567} & \textbf{0.6341} & \textbf{0.6123} \\
        \bottomrule
    \end{tabular}
    }
    \captionsetup{width=\linewidth}
    \vspace{-3mm}
    \label{tab:Zero-shot_accuracy_qwen_3bit}
\end{table}

The hyperparameters used in our experiments are shown in Table \ref{tab:hyperparameters_value}. All experiments are conducted on NVIDIA 4090 GPUs using PyTorch 2.0 \cite{Paszke_2019_PyTorcha} with HuggingFace integration \cite{wolf_2019_huggingface}, to ensure the consistent comparison with baseline approaches. For fair evaluation, we maintain identical experimental settings across all quantization approaches, including calibration data (128 randomly sampled sequences from C4).

\subsection{Experimental Results}

In this section, we present the experimental results in terms of the PPL with 3 benchmarks, the accuracy on zero-shot tasks, and the evaluation of \name{} based on LLM-as-a-judge comparison.

\subsubsection{Perplexity Evaluation}
\label{subsubsec:ppl}
As shown in Table \ref{tab:ppl_llama}, \name{} consistently achieves excellent performance in terms of PPL across 2 LLMs and 3 benchmarks when performing 4-bit and 3-bit quantization on average. To mitigate the influence of evaluation randomness, all PPL results reported in this section are averaged over 3 independent runs with different random seeds for calibration sampling, and we report the mean value across runs. With LLaMA-7B and 4 average bits, \name{} attains the PPLs of 5.81 on WikiText-2 and 10.34 on PTB, outperforming the strongest baseline (AWQ) by 0.02 and 0.08, respectively. While \name{} corresponds to slightly higher (0.01) PPL compared with AWQ and OmniQuant on C4, it still significantly outperforms other baseline approaches (from 0.13 to 3.39). We observe similar results with LLaMA-13B at 4 bits. Under 3-bit quantization, the advantage of \name{} becomes substantially larger: \name{} outperforms all baselines on both models, reducing average PPL by up to 2.37 over GPTQ and 0.40 over AWQ on LLaMA-7B. In addition, \name{} outperforms baseline approaches (from 0.04 to 2.86 in average PPL) on Qwen2.5-7B, Qwen2.5-14B, and Mistral-7B-v0.1 (see Table \ref{tab:ppl_qwen_mistral_4bit} in Appendix \ref{subsubsec:results}).


\subsubsection{Zero-shot Reasoning Evaluation}


As shown in Table \ref{tab:Zero-shot_accuracy_qwen_3bit}, we evaluate zero-shot reasoning using lm-evaluation-harness (see visualized Figure~\ref{fig:radar_3bit} in Appendix for the 3-bit results of Qwen2.5-7B). Compared with RTN and GPTQ, which it outperforms by 6.87\% and 6.69\% in average accuracy respectively, \name{} avoids the severe shrinkage of the radar profile, indicating better preservation of general reasoning ability under aggressive compression. Compared with stronger quantized baselines such as AMQ, AWQ and OmniQuant, \name{} expands the outer boundary on most tasks and remains closer to the FP16 reference, especially on BoolQ, ARC-E, and WinoGrande, demonstrating that layer-adaptive bit-width allocation is critical for preserving model quality under aggressive compression.


\begin{figure}[!t]
\centering
\includegraphics[width=\linewidth]{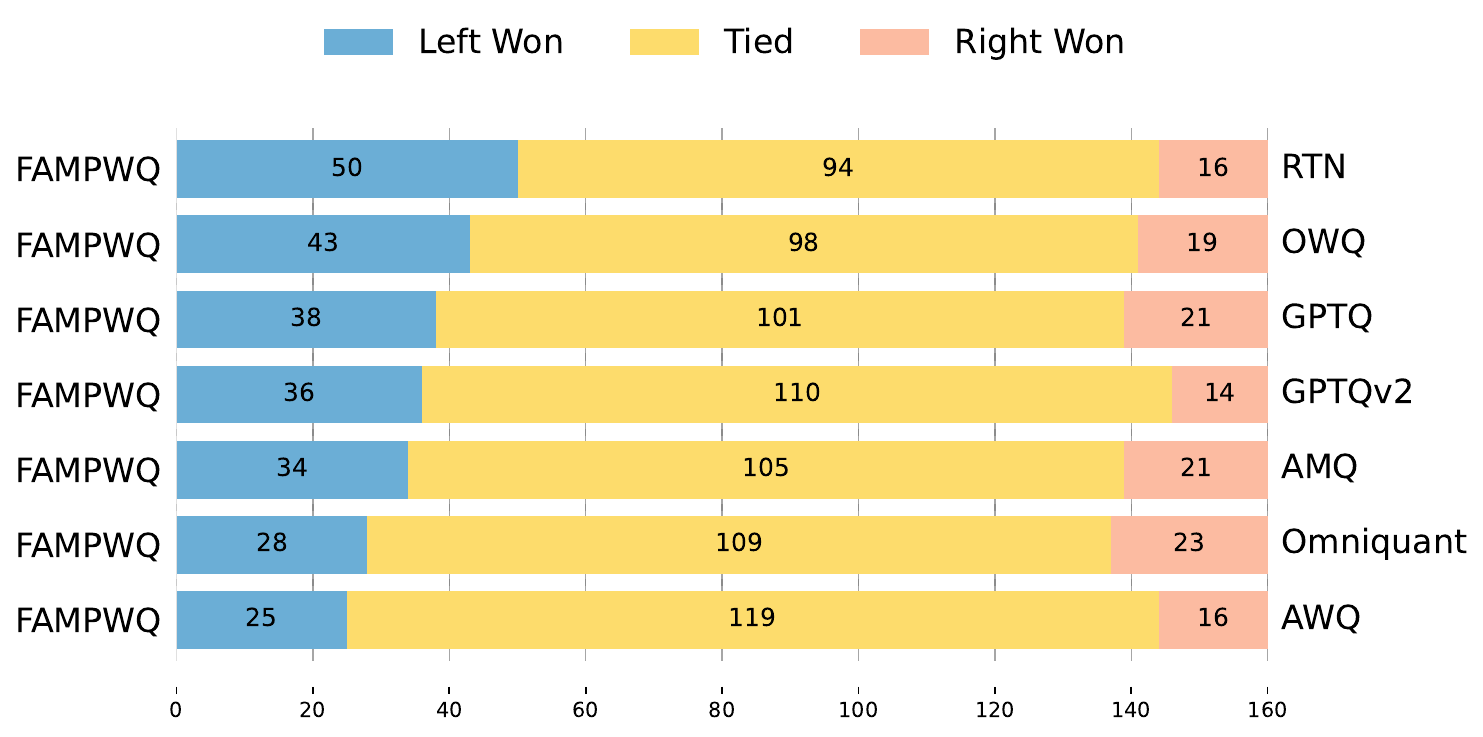}
\captionsetup{width=\linewidth}
\vspace{-8mm}
\caption{LLM-as-a-judge comparison based on GPT-3.5-turbo with 4-bit quantized LLaMA2-13B-chat.}
\label{fig:llm_judge}
\vspace{-7mm}
\end{figure}

\subsubsection{LLM-as-a-judge Evaluation}

To comprehensively evaluate the performance of \name{}, we compare \name{} with baseline approaches based on the quantized versions of the instruction-tuned LLaMA2-13B-chat model exploiting the Vicuna benchmark \cite{chiang_2023_vicuna}. We use GPT-3.5-turbo \cite{gpt3.5turbo} as a judge across 80 diverse questions. We mitigate position bias through bidirectional comparison, which results in 160 trials per comparison. As shown in Figure \ref{fig:llm_judge}, \name{} achieves substantially higher win rates than all baseline approaches (76\% against RTN, 69\% against OWQ, 64\% against GPTQ, 72\% against GPTQv2, 61\% against AWQ and 54\% against OmniQuant), where the win rate excludes tie samples. A two-sided binomial test on the head-to-head trials confirms that the comparison against the strong AWQ baseline is statistically significant ($p<0.05$), reducing the risk that the observed judge preference is caused by evaluation noise. Additional 3-bit Vicuna-Bench results are reported in Appendix \ref{subsubsec:vicuna_3bit}.

\subsubsection{Inference acceleration}

\begin{figure}[t]
\centering
\includegraphics[width=1.0\linewidth]{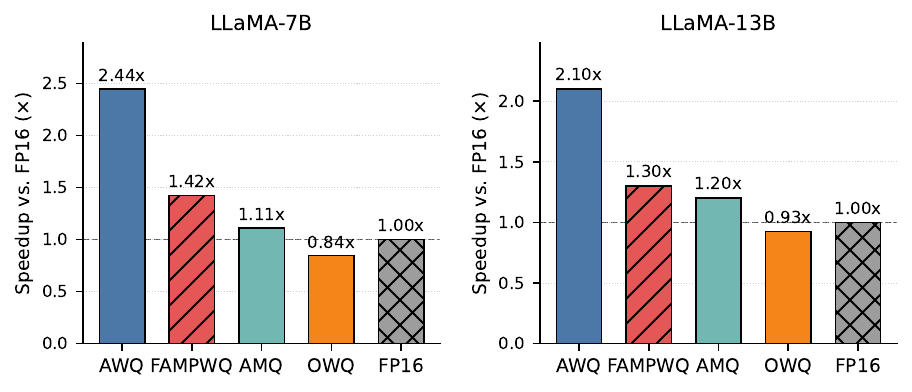}
\captionsetup{width=\linewidth}
\vspace{-6mm}
\caption{Inference speedup over FP16 on NVIDIA 4090 for 3-bit average quantization.}
\label{fig:inference_speed}
\vspace{-2mm}
\end{figure}

As shown in Figure~\ref{fig:inference_speed}, \name{} delivers a clear throughput advantage over FP16 (up to $42\%$) and both intra-layer (OWQ) (up to $69\%$) and activation-guided (AMQ) (up to $28\%$) mixed-precision baselines, while remaining slower than uniform low-bit AWQ due to heterogeneous kernel scheduling. 
AWQ retains the highest absolute throughput ($2.44\times$ on 7B, $2.10\times$ on 13B) by exploiting uniform 4-bit kernels, while \name{} delivers consistently higher accuracy at the same or lower average bit-width as shown in Table~\ref{tab:ppl_llama}.

\subsubsection{Computational Cost}
\label{subsubsec:computational_cost}

\begin{figure}[!t]
\centering
\includegraphics[width=\linewidth]{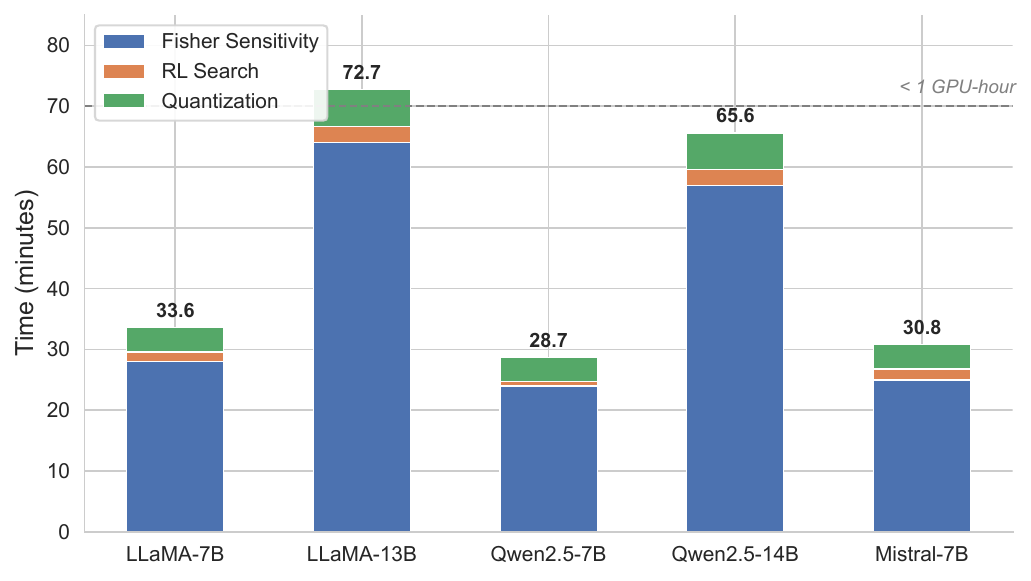}
\captionsetup{width=\linewidth}
\vspace{-4mm}
\caption{Preprocessing time breakdown of \name{} across 5 models. Total cost remains below 1 GPU-hour even for 14B-scale models. Fisher sensitivity computation dominates, while RL search takes only 1--3 minutes.}
\label{fig:computational_cost}
\vspace{-4mm}
\end{figure}

The preprocessing overhead of \name{} consists of three components: Fisher sensitivity computation, RL-based bit-width search, and the quantization itself. As shown in Figure~\ref{fig:computational_cost}, the total preprocessing time remains below 1 GPU-hour for all models tested, including 14B-scale models. Fisher sensitivity computation dominates the cost (24--64 minutes) and scales with model size. The RL search is lightweight ($<$5 minutes on a single GPU), as it operates on a proxy model rather than performing full quantization at each step. The entire preprocessing is a one-time offline cost, amortized across all subsequent inference.

\begin{figure}[!t]
\centering
\includegraphics[width=\linewidth]{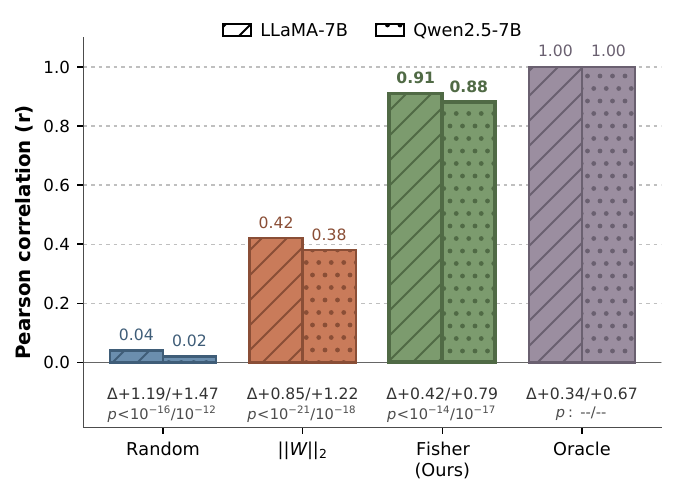}
\captionsetup{width=\linewidth}
\vspace{-8mm}
\caption{Pearson correlation ($r$) between sensitivity metrics and actual quantization degradation (Oracle).}
\label{fig:sensitivity_metric}
\vspace{-4mm}
\end{figure}

\begin{table}[t]
    \centering
    \caption{PPL $\downarrow$ of LLaMA-7B and Qwen2.5-7B on WikiText-2 with diverse perturbation strategies. \textbf{Bold} indicates the lowest PPL. $x\%\delta_1\theta$ refers to the $\delta_1$ strategy with $\epsilon = x\%$. $x\%\delta_2\theta$ denotes the $\delta_2$ strategy with $\beta = x\%$. $x$bit$\delta_3\theta$ represents the $\delta_3$ strategy with $b = x$.}
    \label{tab:Artificial disturbance}
    \vspace{-3mm}
    \renewcommand{\arraystretch}{1}
    \resizebox{0.92\hsize}{!}{
    \begin{tabular}{l|ccc|ccc|cc}
        \toprule
        \textbf{Perturbation Type} & $1\%  \delta_1 \theta$ & $10\%  \delta_1 \theta$ & $20\%  \delta_1 \theta$ & $1\%  \delta_2 \theta$ &$10\%  \delta_2 \theta$ & $20\%  \delta_2 \theta$ &$4 \text{bit}  \delta_3 \theta$ & $8 \text{bit}  \delta_3 \theta$   \\
        \midrule
        LLaMA-7B & 6.61 & 6.57 & 6.68 & 6.62 & 6.61 & 6.72 & \textbf{6.49} & 6.53 \\
        Qwen2.5-7B & 8.42 & 8.37 & 8.83 & 8.74 & 8.36 & 8.40 &\textbf{8.27} & 8.33\\
        \bottomrule
    \end{tabular}
    }
\vspace{-4mm}
\end{table}

\subsection{Ablation Study}

In this section, we analyze the impact of diverse sensitivity measurement methods and the comparison of diverse bit-width allocation methods.

\subsubsection{Artificial Perturbation}

We compare $\delta\theta_l$ in Eq.~\ref{eq:delta3} against two generic alternatives: magnitude-proportional uniform noise ($\delta_1$) and Bernoulli-masked weight-proportional noise ($\delta_2$) (see Appendix~\ref{AppSubSec:perturbation_ablation} for details). As shown in Table~\ref{tab:Artificial disturbance}, the quantization perturbation form at $b{=}4$ yields the lowest PPL on both models, beating $\delta_1$ by up to 0.56 and $\delta_2$ by up to 0.23. Only $\delta_3$ matches the actual $b$-bit rounding perturbation in both direction and magnitude; $\delta_1$ and 
$\delta_2$ are agnostic to the target bit-width and therefore reflect only generic parameter importance.

\subsubsection{Bit-width Allocation Strategy}

We compare our PPO-based method with four alternative methods: greedy search, Bayesian optimization, simulated annealing, and a genetic algorithm. As shown in Table \ref{tab:allStrategy}, the RL-based adaptive allocation strategy achieves substantially lower average PPL than these alternatives (up to 1.50 lower than Greedy, 1.33 lower than Bayesian optimization, 1.90 lower than simulated annealing, and 0.45 lower than the genetic algorithm), revealing the superb performance of our allocation method.

\subsubsection{Sensitivity Metric Comparison}
\label{subsec:sensivity}

\begin{table}[t]
    \centering
    \caption{{PPL $\downarrow$ with different bit-width allocation methods. \textbf{Bold} indicates the lowest PPL. }}
    \label{tab:allStrategy}
    \vspace{-3mm}
    \renewcommand{\arraystretch}{1}
    \resizebox{\hsize}{!}{
    \begin{tabular}{l|c|cccc|cccc}
        \toprule
        \textbf{Model} & \multicolumn{1}{c}{}& \multicolumn{4}{c|}{\textbf{LLaMA-7B}} &
        \multicolumn{4}{c}{\textbf{Qwen2.5-7B}} \\
        \midrule
        \textbf{Strategy} & \multicolumn{1}{c|}{Avg bit} & Wiki2 & PTB & C4 & Avg PPL & Wiki2 & PTB & C4 & Avg PPL \\
        \midrule
        FP16 & 16 & 5.68 & 10.11 & 7.34 &7.71 & 6.84 & 12.79 & 11.88 & 10.50\\
        \midrule
        Greedy & 3 & 6.90 & 12.12 & 9.22 & 9.41 & 8.81 & 16.05 & 14.16 & 13.67 \\
        Bayesian & 3 & 7.29 & 12.57 & 9.86 & 9.91 & 8.37 & 15.24 & 13.78 & 12.46 \\
        Annealing & 3 & 7.54 & 13.85 & 10.06 & 10.48 & 8.86 & 16.59 & 14.65 & 13.37 \\
        Genetic & 3 & 6.59 & 11.71 & 8.79 & 9.03 & 8.27 & 15.26 & 13.64 & 12.39 \\
        RL (Ours) & 3 & \textbf{6.35} & \textbf{11.33} & \textbf{8.07} & \textbf{8.58} & \textbf{8.08} & \textbf{14.99} & \textbf{13.45} & \textbf{12.17} \\
        \bottomrule
    \end{tabular}
    }
\vspace{-4mm}
\end{table}

We compare the FIM-based sensitivity metric against random allocation, weight magnitude ($\|W\|_2$), and Oracle sensitivity (actual per-layer PPL increase). As shown in Figure~\ref{fig:sensitivity_metric}, our FIM-based sensitivity metric achieves a Pearson correlation of $r{=}0.91, p<10^{-14}$ on LLaMA-7B and $r{=}0.88, p < 10^{-17}$ on Qwen2.5-7B with Oracle sensitivity.
This significant correlation directly leads to better quantization performance: at 3.5-bit average, the FIM-based metric limits $\Delta$PPL to +0.42, while weight magnitude yields +0.85 and random allocation yields +1.19 (see detals in Appendix Table~\ref{tab:sensitivity_comparison}). 

\subsubsection{$\alpha$ Sensitivity Analysis}

\begin{figure}[!t]
\centering
\includegraphics[width=\linewidth]{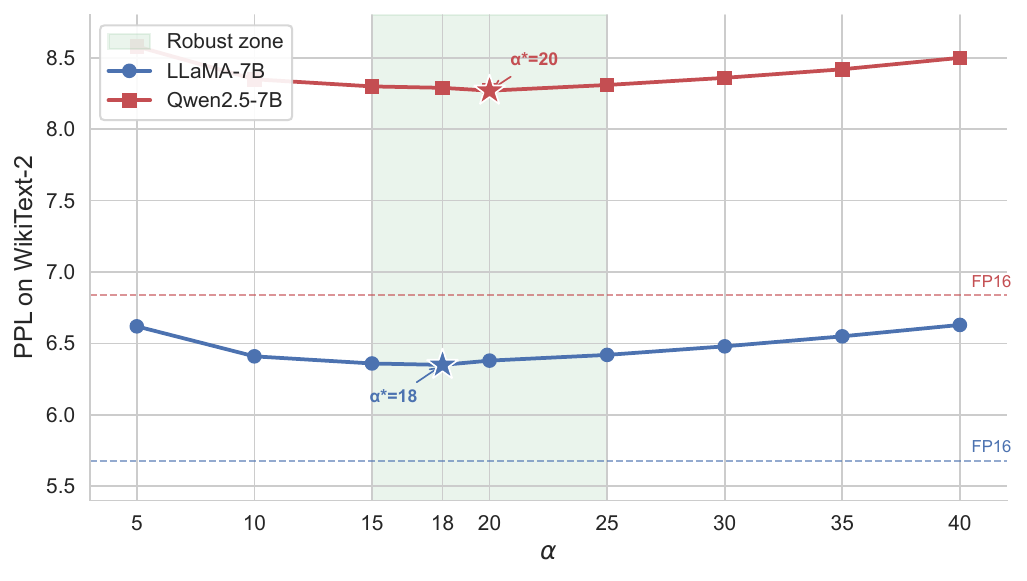}
\captionsetup{width=\linewidth}
\vspace{-9mm}
\caption{PPL on WikiText-2 as a function of decay rate $\alpha$ for LLaMA-7B and Qwen2.5-7B under 3-bit average quantization. Stars mark the optimal $\alpha$ for each model. The shaded green region indicates the robust zone ($\alpha \in [15, 25]$) where PPL variation is minimal ($<$0.3). Dashed lines show FP16 baselines.}
\label{fig:alpha_sensitivity_curve}
\vspace{-6mm}
\end{figure}

As shown in Figure~\ref{fig:alpha_sensitivity_curve}, the decay rate parameter $\alpha$ exhibits a broad optimal region. For LLaMA-7B, the optimal $\alpha{=}18$ yields a PPL of 6.35, while any $\alpha \in [15, 25]$ produces PPL within 0.09 of the optimum. For Qwen2.5-7B, $\alpha{=}20$ is optimal (PPL 8.27), with $<$0.06 variation across the robust zone. This robustness to $\alpha$ simplifies hyperparameter selection and confirms that the exponential decay model in Formula~\ref{eq:accuracy_loss2single_layer} is a stable approximation.




\section{Conclusion}

In this work, we propose a novel Fisher information-based Adaptive Mixed Precision Weight Quantization approach, i.e., \name{}. \name{} introduces a novel perturbation-based Fisher Information metric to capture layer-wise quantization-specific sensitivity. In addition, \name{} couples the metric with a new PPO-based allocation method to efficiently generate an adaptive bit-width allocation strategy with superb performance. Extensive experiments on 7 models and 5 benchmarks demonstrate that \name{} outperforms 7 baselines in PPL (up to 3.39 smaller), accuracy (up to 6.87\% higher), and LLM-as-a-judge comparison (up to 76\% win rate), with particularly strong advantages at the 3-bit compression frontier.

\section*{Limitations}
Several limitations remain for \name{}. First, mixed-precision quantization can reduce inference throughput while heterogeneous bit-widths are less compatible with optimized uniform-precision kernels. Second, \name{} currently targets weight-only quantization (W$x$A16), leaving joint weight-activation quantization to future work. Third, our experiments focus on dense Transformer models, and effectiveness on Mixture-of-Experts architectures remains untested. Finally, Fisher sensitivity estimation is a one-time offline cost but remains the dominant preprocessing component, motivating lighter sensitivity proxies.

\section*{Ethical Considerations}
We focus on the quantization methods for LLMs in this paper. Our work does not involve the collection of new data, human subjects, or user-generated content. All experiments are conducted on publicly available datasets. We do not identify new ethical concerns arising from this work. The proposed method is model-agnostic and does not introduce new risks related to bias, privacy, or fairness beyond the underlying models and datasets. Nevertheless, we acknowledge that compressed LLMs may be subject to potential risk if deployed without appropriate safeguards. We exploited LLMs to polish the expression of this paper.

\section*{Acknowledgements}
This work was partially (for Juncheng Jia) supported by the Priority Academic Program Development of Jiangsu Higher Education Institutions, Suzhou Frontier Science and Technology Program (Project SYG202310).

\bibliography{custom}

\clearpage
\newpage

\appendix
\label{sec:appendix}
\setcounter{figure}{0}
\renewcommand*{\thefigure}{A.\arabic{figure}}

\setcounter{table}{0}
\renewcommand*{\thetable}{A.\arabic{table}}

\setcounter{equation}{0}
\renewcommand\theequation{A.\arabic{equation}}

\begin{table*}[t!] 
    \centering
    \vspace{-4mm}
    \caption{Summary of main notations}
    \label{tab:Symbol Table}
    \vspace{-4mm}
    \renewcommand{\arraystretch}{1}
    \resizebox{\hsize}{!}{
    \begin{tabular}{cc}
    \toprule
        Symbols & Description\\
    \midrule
        $\theta;\theta_l$ & LLM Parameter; parameters in Layer $l$.\\
        $F(\cdot);\hat{F}(\cdot);\mathcal{F}(\cdot)$ & Fisher Information Matrix (FIM); empirical FIM; the diagonal vector of FIM. \\
        $p_{\theta_l}(y|x)$  & The probability density function of the inference with $\theta_l$.  \\
        $\nabla_{\theta_l}\log p_{\theta_l}(y|x)$ & The first-order derivative of $\theta_l$, which is calculated by the gradient.\\
        $\mathcal{D};\mathcal{L}$  &  Validation dataset; quantization layer set. \\
        $\mathcal{S};s_l$ & The set of sensitivity of all the layers; the sensitivity of Layer $l$. \\
        $p_l; R$ & The number of parameters in Layer $l$; target compression ratio.\\
        $\mathcal{B};B$ & The set of candidate bit-widths; original (full-precision) bit-width. \\
        $\mathcal{Q};q$ & The quantization bit-width allocation strategy; bit-width allocation action.\\
        $\Delta Acc$ & The accuracy degradation incurred by quantizing an individual layer.\\
        $\alpha$ & Positive constant decay rate.\\
        $ \delta\theta_l; \theta^{\text{pert}}_l$ & Noise to $\theta_l$; perturbed layer parameters. \\
        $\nabla_{\theta_l} \ell(x|\theta)$ & The first-order derivative of the LLM.\\
        $\mathcal{L}(\cdot);\psi(\cdot);P$ & Loss function of the bit-width allocation problem; compression target loss; penalty or reward value.\\
        $\theta_t;\phi_t$ & The parameters of the actor network at Step $t$; the parameters of the critic network at Step $t$.\\
        $V_{\phi_t};A_t$ & The scalar value of the critic network at Step $t$; Temporal-Difference (TD) advantage at Step $t$. \\
        $\eta_\theta;\eta_\phi$ & The learning rate of the actor network; the learning rate of the critic network.\\
        $\mathfrak{r}_t(\theta);\mathfrak{c}_t(\theta)$ & Policy-dependent reward for conservative strategy adaptation; the clip reward for limiting update range. \\ 
    \bottomrule     
    \end{tabular}
    }
\end{table*}

\section{Appendix}
\subsection{Explanation of Notations}
The meanings of the notations in this paper are summarized in Table \ref{tab:Symbol Table}.

\subsection{Preliminary}

In this section, we present the quantization preliminary and introduce Fisher information calculation. 

\subsubsection{Quantization Preliminary}
\label{AppSubSec:Quant}

The quantization process can be classified into uniform and non-uniform \cite{wang2024model}. 
The uniform quantization uses uniform and finite intervals (e.g., $2^b$ intervals for $b$-bit integer) to represent the original values. In contrast, non-uniform quantization utilizes non-uniformly spaced intervals, and the length of intervals can vary. Given a weight tensor $\mathbf{W}$ in a LLM, the quantization and de-quantization process can be defined as Formula \ref{eq:quant}.
\begin{equation}\label{eq:quant}
  \begin{aligned}
\mathbf{W}^{Q}=Q(\mathbf{W}),\quad \mathbf{\widetilde{W}}=Q^{-1}(\mathbf{W}^{Q}),
  \end{aligned}
\end{equation}
where $\mathbf{W}^{Q}$ is the quantized tensor, and $\mathbf{\widetilde{W}}$ is the recovered tensor. The quantization function $Q(\cdot)$ of a uniform quantization approach is defined as a rounding-to-nearest operation over the scaled input calculated in Formula \ref{eq:quantOperation}.
\begin{equation}\label{eq:quantOperation}
  \begin{aligned}
    Q_{\text{uni}}(\mathbf{W})=\text{clip}\left(\left \lfloor \frac{\mathbf{W}}{\alpha} \right \rceil+z;0,2^b-1 \right),
  \end{aligned}
\end{equation}
where $b\in \mathbb{N} $ is the bit-width, $\alpha \in \mathbb{R} $ is the scale factor, $z\in \mathbb{N} $ is zero-point or offset value, $\left \lfloor \cdot \right \rceil$ denotes the round-to-nearest-integer operator, and $clip(x,min_{value},max_{value})$ represents a clip function of the input $x$ with the minimum value and the maximum value. The corresponding de-quantization function is defined in Formula \ref{eq:dequant}.
\begin{equation}\label{eq:dequant}
  \begin{aligned}
    Q_{\text{uni}}^{-1}(\mathbf{W})=(\mathbf{W}^{Q}-z)\cdot \alpha.
  \end{aligned}
\end{equation}

Uniform quantization can be either symmetric or asymmetric according to the sign of the mapping space. In this paper, we use the symmetric uniform  quantization approach. The symmetric quantization restricts the zero-point to 0 as defined in Formula \ref{eq:symQuant}.
\begin{equation}\label{eq:symQuant}
  \begin{aligned}
    Q_{\text{uni\_sym}}(\mathbf{W})=\text{clip}\left(\left \lfloor \frac{\mathbf{W}}{\alpha}\right \rceil;-2^{b-1},2^{b-1}-1 \right)
  \end{aligned}
\end{equation}

Quantization approaches optimize the global loss as defined in Formula \ref{eq:quantization_loss}.
\begin{equation}\label{eq:quantization_loss}
  \begin{aligned}
    \operatorname*{argmin}_{\mathbf{W}^Q} E & = \operatorname*{argmin}_{\mathbf{W}^Q} \left \| \mathbf{W}X-\mathbf{W}^QX \right \|^2_2,
  \end{aligned}
\end{equation}
where $X$ is the input of the corresponding layer of the LLM.


\subsubsection{Fisher Information}
\label{AppSubSubSec:FIM}

Fisher information quantifies the amount of information that observable data carries about the unknown parameters of a probabilistic model \cite{Ly_2017_Tutorial,Zheng_2024_MixLLMa}. Fisher information can be used to evaluate parameter importance in neural networks by quantifying how sensitive the model output is to the noises of each parameter $\theta$. We can denote the Fisher Information Matrix (FIM) by the expectation of the outer product of score vectors \cite{Amari_1998_Natural} as defined in Formula \ref{eq:fisher-info}.
\begin{equation}\label{eq:fisher-info}
F(\theta_l):=
\textstyle \mathbb{E}_{p_\theta(y|x)}\left [ \nabla_{\theta_l}\log p_{\theta_l}(y|x)\nabla_{\theta_l}\log p_{\theta_l}(y|x)^\top  \right ],
\end{equation}
where $p_{\theta_l}(y|x)$ represents the probability density function of the inference with LLM and parameters $\theta_l$ at Layer $l$, $\nabla_{\theta_l}\log p_{\theta_l}(y|x)$ denotes the first-order derivative of the LLM, which is calculated via the gradient. In practice, we use the empirical FIM to approximate the expected one \cite{kunstner2019limitations} as shown in Formula \ref{eq:empirical}.
\begin{equation}\label{eq:empirical}
\hat{F}(\theta_l) = \frac{1}{N}\sum_{n=1}^N \nabla_{\theta_l}\log p_{\theta_l}(y_n|x_n)\nabla_{\theta_l}\log p_{\theta_l}(y_n|x_n)^\top,
\end{equation}
where $N$ represents the number of samples in the validation dataset $\mathcal{D}$.

\begin{figure*}[!t]
\centering
\includegraphics[width=0.8\linewidth]{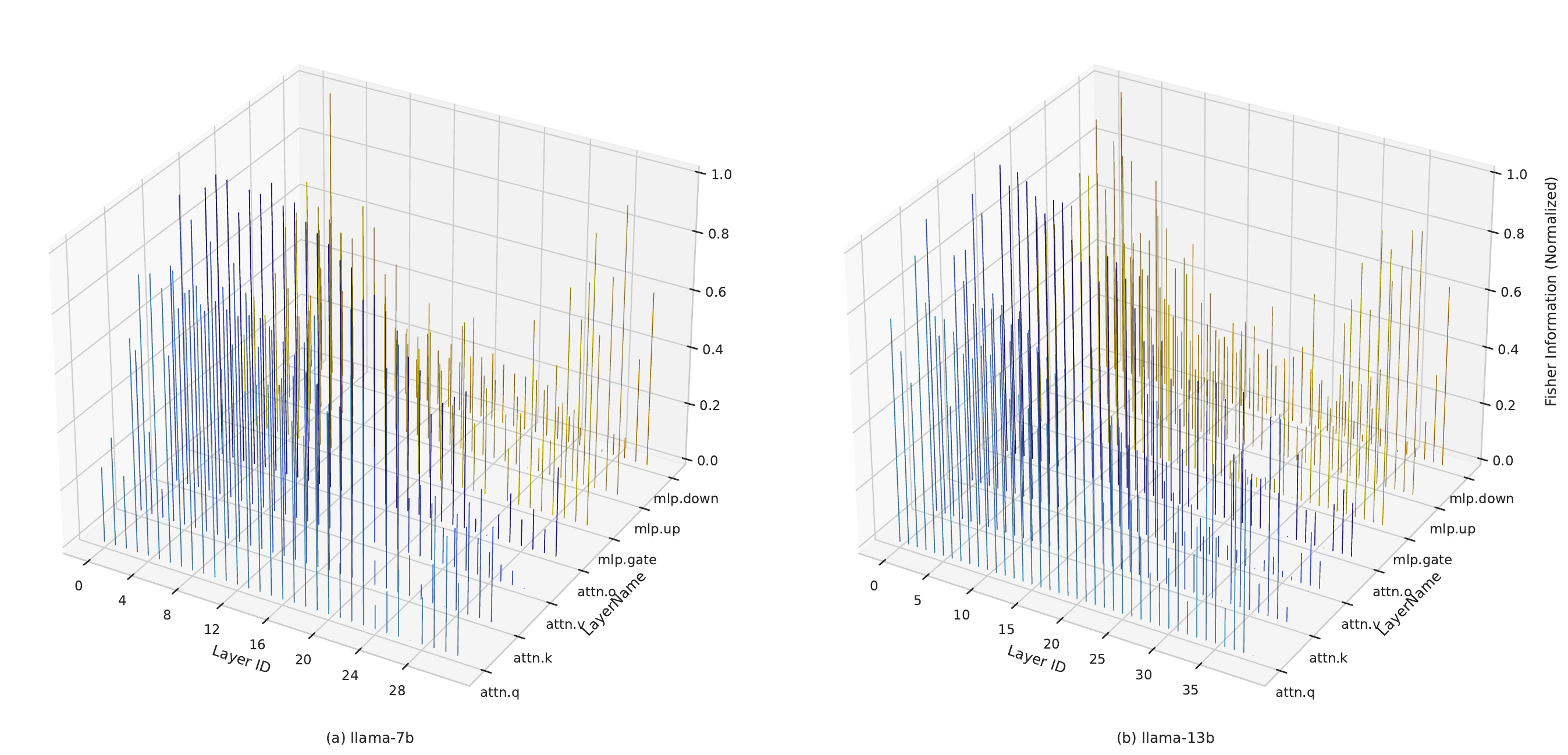}
\captionsetup{width=\linewidth}
\vspace{-4mm}
\caption{Visualization of FIM for LLaMA-7B and LLaMA-13B.}
\label{fig:fisher}
\vspace{-4mm}
\end{figure*}

FIM captures the essential influence of parameters on the likelihood function, where larger FIM values indicate more influential parameters should be preserved for inference. FIM can be used to evaluate the importance of layer-wise parameters \cite{Tu_2016_Ranking}, so as to preserve accuracy while compressing LLMs \cite{Singh_2020_WoodFishera,Liu_2021_Group}. While calculating FIM is computationally expensive with large gradient matrices, we use the diagonal vector of FIM to represent the FIM \cite{Liu_2024_Fishera} as defined in Formula \ref{eq:FIMrank}.
\begin{equation}\label{eq:FIMrank}
\mathcal{F}(\theta_l) = \text{diag}(I_{|\nabla_{\theta_l} \log p_{\theta_l}(y|x)|} \odot \hat{F}(\theta_l)), 
\end{equation}
where $\text{diag}(\cdot)$ represents the diagonal vector of a matrix, $I_{|\nabla_{\theta_l} \log p_{\theta_l}(y|x)|}$ is the identity matrix with the same size of the gradient matrix, and $\odot$ is element-wise multiplication.

\subsection{Pearson Correlation Analysis}

In order to verify the correlation between FIM and accuracy degradation, we calculate the Pearson correlation coefficients \cite{benesty2009noise} between the layer-wise quantization sensitivity and the increase in perplexity (PPL) with perturbations ($\delta_3$ quantization-simulating perturbations at $b{=}4$) in each layer. As shown in Table \ref{tab:pearson_analysis}, the perturbations and the increase in PPL have a significant positive correlation with $R > 0.5$ and $P < 0.0001$, indicating that FIM is a reliable proxy for performance degradation.

\begin{table}[t]
    \centering
    \caption{Pearson correlation coefficients between FIM and the increase in PPL among different types of LLaMA-7B layers. $R$ represents the related correlation coefficient and $P$ refers to the Pearson value.}
    \label{tab:pearson_analysis}
    \vspace{-4mm}
    \renewcommand{\arraystretch}{1}
    \resizebox{0.85\hsize}{!}{
    \begin{tabular}{l|c|c}
        \toprule
        \textbf{layer name} & $R$ & $P$ \\
        \midrule
        attn.q &0.884251&  1.732190 $\times 10^{-43}$\\
        attn.k &0.917151&  3.526569 $\times 10^{-52}$\\
        attn.v &0.761516&  1.728182 $\times 10^{-25}$\\
        attn.o &0.811218&  3.781759 $\times 10^{-31}$\\
        mlp.gate &0.717476&  1.608838 $\times 10^{-21}$\\
        mlp.up   &0.699258&  4.318466 $\times 10^{-20}$\\
        mlp.down &0.612742&  1.509229 $\times 10^{-14}$\\
        \bottomrule
    \end{tabular}
    }
    \vspace{-4mm}
\end{table}

\begin{figure*}[!t]
\centering
\includegraphics[width=0.8\linewidth]{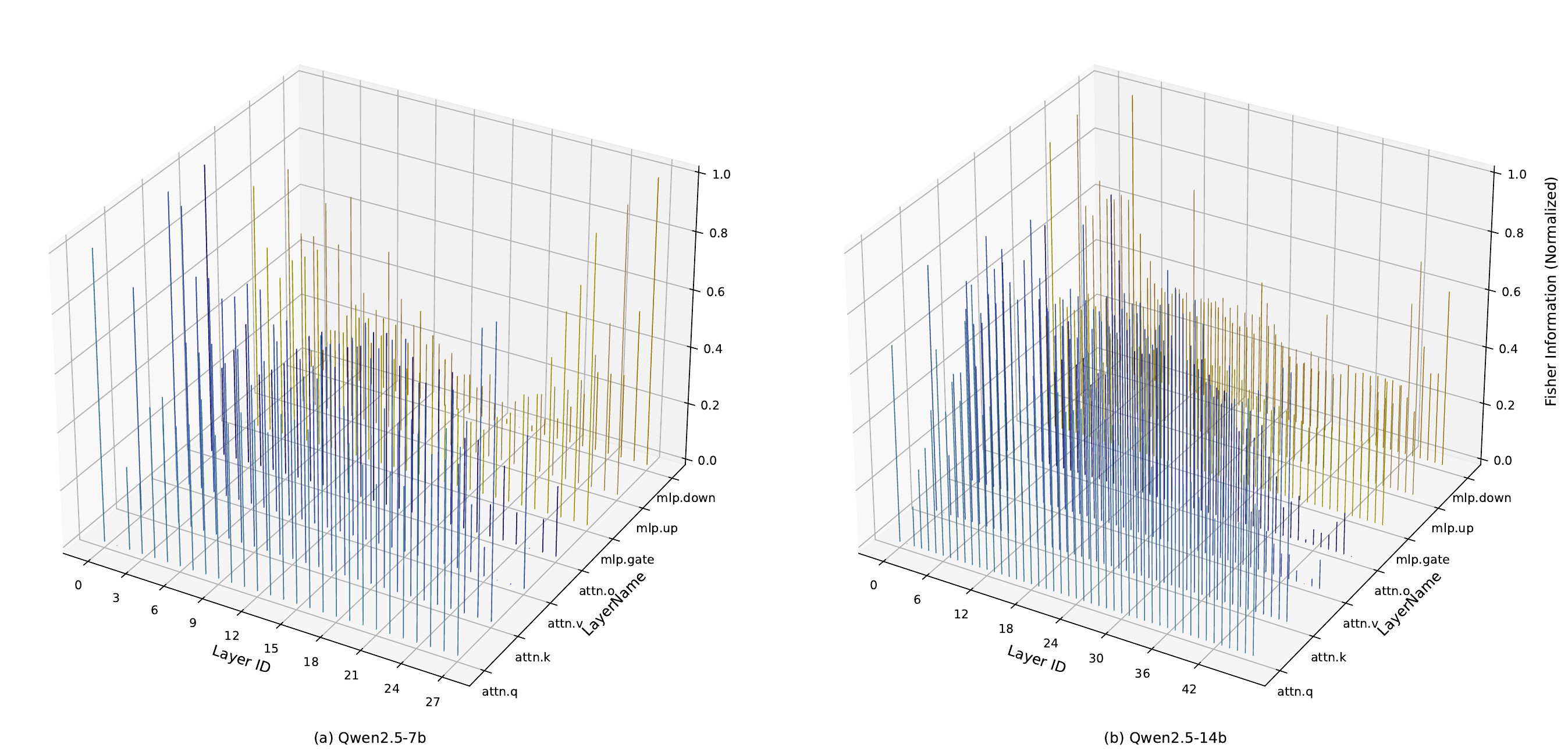}
\captionsetup{width=\linewidth}
\vspace{-4mm}
\caption{Visualization of FIM for Qwen2.5-7B and Qwen2.5-14B.}
\label{fig:fisher of qwen}
\vspace{-4mm}
\end{figure*}

\subsection{Perturbation Strategy Ablation}
\label{AppSubSec:perturbation_ablation}

In Section~\ref{subse:fisherinfo} we adopt the quantization perturbation $\delta\theta_l = Q_b(\theta_l) - \theta_l$ as the canonical choice, since $\theta_l + \delta\theta_l = Q_b(\theta_l)$ and it therefore matches the exact perturbation that $b$-bit quantization injects. For completeness, here we provide the full mathematical definitions of two alternative perturbation forms used in the ablation in Section~5.4.1:
\begin{itemize}
    \item Magnitude-proportional uniform noise: $\delta_1\theta_l = \epsilon \cdot \mu_l$, where $\mu_l = \mathbb{E}[\theta_l]$ and $\epsilon \sim \mathcal{U}(-1,1)$. This form scales uniform noise by each layer's mean magnitude, preserving relative scale differences across layers but \emph{not} the structure of the quantization perturbation.
    \item Bernoulli-masked weight-proportional noise: $\delta_2\theta_l = \beta \cdot \theta_l \odot \mathbf{m}$, with $\beta \in [0,1]$ and $\mathbf{m} \sim \text{Bernoulli}(0.5)$. This form applies sparse, weight-proportional noise via a random mask. The fixed ratio $\beta$ ensures consistent perturbation intensity but, like $\delta_1$, is agnostic to the target bit-width $b$.
\end{itemize}
The corresponding empirical comparison is reported in the main text (Table~\ref{tab:Artificial disturbance}); both alternatives are dominated by the quantization perturbation $\delta_3$ at $b{=}4$.

\subsection{Visualization of Sensitivity}
As shown in Figures \ref{fig:fisher} and \ref{fig:fisher of qwen}, the sensitivity (FIM) varies substantially across layers. Input and output-adjacent layers tend to be more sensitive because they shape low-level feature encoding and final predictions, while intermediate layers show more heterogeneous behavior. \name{} exploits this diversity by preserving high-sensitivity layers and compressing low-sensitivity layers more aggressively.

\begin{algorithm}[!t]
\small

\caption{Reinforcement Learning (RL)-based Network Training}

\textbf{Input:} \\
\hspace{1em} $E$:  The number of training epochs \\
\hspace{1em} $L$:  The number of layers in a LLM \\
\textbf{Output:} \\
\hspace{1em} $\theta;\phi$: Parameters of the pre-trained actor and critic network \\ 

\begin{algorithmic}[1]
\STATE $\theta_0$;$\phi_0 \gets$ \text{Randomly initialize the actor and critic network} 
\STATE $\mathcal{Q}_0 \gets [\max(\mathcal{B})]_{L}$ 
\FOR{Epoch $e = 1$ to $E$}
    \FOR{Step $t = 1$ to $L$}
        \STATE $q^t_t \gets$ Generate the bit-width for Layer $t$
        \STATE $\mathcal{Q}_{t} \gets \text{Update } \mathcal{Q}_{t-1}$ \text{by replacing} $q_t^{t-1}$ by $q^t_t$ 
        \STATE $\mathcal{L}(\mathcal{Q}_{t-1})
        \gets$ Calculate the loss according to Formula~\ref{eq:rl_objective}
        \STATE $A_t \gets$ \text{Calculate according to Formula~\ref{eqn:advantage}}
        \STATE $\mathfrak{r}_t(\theta_{t-1}) \gets$ \text{Calculate} $\mathfrak{r}_t(\theta_{t-1})$ \text{according to Formula~\ref{eq:policy_reward}}
        \STATE $\mathfrak{c}_t(\theta_{t-1}) \gets$ \text{Calculate} $\mathfrak{c}_t(\theta_{t-1})$ \text{with} $\mathfrak{r}_t(\theta_{t-1})$ \text{and} $A_t$ \text{according to Formula~\ref{eq:clip_reward}}
        \STATE $\theta_t \gets$ \text{Update} $\theta_{t-1}$ \text{with} $A_t$, $\mathfrak{r}_t(\theta_{t-1})$ \text{and} $\mathfrak{c}_t(\theta_{t-1})$ \text{according to Formula~\ref{eq:upgrade_theta}}
        \STATE $\phi_t \gets$ \text{Update} $\phi_{t-1}$ \text{with} $A_t$ \text{according to Formula~\ref{eqn:upgrade_phi}}
    \ENDFOR
\ENDFOR
\end{algorithmic}
\label{alg:rl_network_training}
\end{algorithm}

\begin{table}[t] 
    \centering
    \caption{Hyperparameter values. $\epsilon$ refers to the hyperparameter in Formula~\ref{eq:clip_reward}. For different models, recommendation $\alpha$ is given.}
    \label{tab:hyperparameters_value}
    \vspace{-4mm}
    \resizebox{0.6\hsize}{!}{
    \begin{tabular}{cc}
    \toprule
        Hyperparameters & Value \\
    \midrule
        $\eta_\theta;\eta_\phi$ & $0.0001$; $0.0003$ \\
        $\epsilon$ & $0.2$ \\
        $\gamma$ & $0.99$ \\
        $P_{\text{penalty}}$ & 10000 \\
        $P_{\text{reward}}$ & -1 \\
        $\alpha$ (LLaMA-7B) & $18$\\
        $\alpha$ (LLaMA-13B) & $15$\\
        $\alpha$ (LLaMA2-7B-chat) & $20$ \\
        $\alpha$ (LLaMA2-13B-chat) & $20$ \\
        $\alpha$ (Qwen2.5-7B) & $20$ \\
        $\alpha$ (Qwen2.5-14B) & $20$ \\
        $\alpha$ (Mistral-7B-v0.1) & $30$ \\
    \bottomrule     
    \end{tabular}
    }
\end{table}

\begin{table}[t]
    \centering
    \vspace{-4mm}
    \caption{PPL of Qwen2.5-7B and LLaMA-7B with quantization of 4-bit and 5-bit and diverse benchmarks (Wiki2, PTB, C4). ``4-bit'' represents that all the layers are quantized to 4-bit. Other rows report the PPL when the corresponding layers are quantized to 5-bit while other layers remain at 4-bit. \textbf{Bold} indicates the lowest PPL. \underline{underlined} indicates the second-highest.}
    \label{tab:Different_strategy}
    \vspace{-4mm}
    \resizebox{\hsize}{!}{
    \begin{tabular}{l|ccc|ccc}
        \toprule
        \textbf{Model} & \multicolumn{3}{c|}{\textbf{Qwen2.5-7B}} &
        \multicolumn{3}{c}{\textbf{LLaMA-7B}} \\
        \midrule
        \textbf{Layer name} & Wiki2 & PTB & C4 & Wiki2 & PTB & C4 \\
        \midrule
        4-bit & 7.094 & 13.218 & 12.228 & 5.834 & 10.420 & 7.528\\
        \midrule
        attn.q & 7.094 & 13.205 & 12.222 & 5.831 & 10.414 & 7.520\\
        attn.k & 7.086 & 13.212 & 12.215 & 5.835 & 10.413 & 7.521\\
        attn.v & 7.072 & 13.170 & 12.193 & 5.783 & 10.367 & \underline{7.487}\\
        attn.o & 7.078 & 13.198 & 12.209 & 5.821 & 10.384 & 7.512\\
        mlp.gate & 7.058 & 13.170 & 12.182 & 5.822 & 10.391 & 7.495\\
        mlp.up & \underline{7.047} & \underline{13.125} & \underline{12.167} & \textbf{5.814} & \textbf{10.350} & 7.493\\
        mlp.down & \textbf{7.041} & \textbf{13.103} & \textbf{12.131} & \underline{5.816} & \underline{10.383} & \textbf{7.484}\\
        \bottomrule
    \end{tabular}
    }
\end{table}

\begin{table}[!ht]
    \centering
    \vspace{-4mm}
    \caption{{PPL of \name{} with varying average quantization bit-widths.}}
    \label{tab:Adaptive Storage}
    \vspace{-4mm}
    \renewcommand{\arraystretch}{1}
    \resizebox{0.8\hsize}{!}{
    \begin{tabular}{c|ccc|ccc}
        \toprule
        \textbf{Model} & \multicolumn{3}{c|}{\textbf{LLaMA-7B}} &
        \multicolumn{3}{c}{\textbf{Qwen2.5-7B}} \\
        \midrule
        \textbf{Avg bit} & Wiki2 & PTB & C4 & Wiki2 & PTB & C4 \\
        \midrule
        16 & 5.68 & 10.11 & 7.34 & 6.84 & 12.79 & 11.88 \\
        \midrule
        3.1 & 6.46 & 11.55 & 8.45 & 8.08 & 14.99 & 13.45 \\
        3.2 & 6.34 & 11.29 & 8.32 & 7.94 & 14.76 & 13.28 \\
        3.3 & 6.22 & 11.04 & 8.17 & 7.84 & 14.60 & 13.15 \\
        3.4 & 6.17 & 10.89 & 8.06 & 7.75 & 14.42 & 13.05 \\
        3.5 & 6.10 & 10.79 & 7.96 & 7.63 & 14.23 & 12.92 \\
        3.6 & 6.04 & 10.65 & 7.85 & 7.55 & 14.07 & 12.78 \\
        3.7 & 5.98 & 10.59 & 7.75 & 7.46 & 13.97 & 12.68 \\
        3.8 & 5.94 & 10.49 & 7.63 & 7.36 & 13.79 & 12.57 \\
        3.9 & 5.85 & 10.40 & 7.57 & 7.24 & 13.52 & 12.42 \\
        4.1 & 5.79 & 10.31 & 7.49 & 7.06 & 13.14 & 12.16 \\
        4.2 & 5.77 & 10.28 & 7.47 & 7.04 & 13.13 & 12.13 \\
        4.3 & 5.75 & 10.27 & 7.45 & 7.02 & 13.09 & 12.11 \\
        4.4 & 5.75 & 10.26 & 7.44 & 7.01 & 13.07 & 12.09 \\
        4.5 & 5.74 & 10.23 & 7.43 & 6.99 & 13.05 & 12.08 \\
        
        \bottomrule
    \end{tabular}
    }
\end{table}

\subsection{Reinforcement Learning (RL)-based Network Training}
\label{AppSubSec:rlTraining}

\begin{table}[t]
    \centering
    \vspace{-4mm}
    \caption{Quantization time with NVIDIA 4090 GPU.}
    \label{tab:time consumption}
    \vspace{-4mm}
    \renewcommand{\arraystretch}{1}
    \resizebox{\hsize}{!}{
    \begin{tabular}{l|ccccccc}
        \toprule
        \textbf{Quant Method} & RTN & GPTQ & GPTQv2 & OWQ & OmniQuant & AWQ & \name{}   \\
        \midrule
        LLaMA-7B & 10s & 369s & 537s & 349s & 600s & 129s & 240s\\
        LLaMA-13B & 12s & 619s & 988s & 598s & 1125s & 240s & 388s\\
        \bottomrule
    \end{tabular}
    }
\end{table}

\begin{table}[t]
    \centering
    \vspace{-4mm}
    \caption{\name{} preprocessing time on NVIDIA 4090 GPUs. Sensitivity is computed using the listed number of GPUs, while RL-based bit-width search is performed on a single GPU.}
    \label{tab:preprocessing_time}
    \vspace{-4mm}
    \renewcommand{\arraystretch}{1}
    \resizebox{\hsize}{!}{
    \begin{tabular}{l|c|c}
        \toprule
        \textbf{Model (\#GPUs)} & Sensitivity calculation (min) & Bit width optimization search (s)   \\
        \midrule
        LLaMA-7B (1) & 28 & 97\\
        LLaMA-13B (2) & 64 & 159\\
        LLaMA2-7B-chat (1) & 28 & 95\\
        LLaMA2-13B-chat (2) & 62 & 148\\
        Qwen2.5-7B (1) & 24 & 43\\
        Qwen2.5-14B (2) & 57 & 156\\
        Mistral-7B-v0.1 (1) & 25 & 110\\
        \bottomrule
    \end{tabular}
    }
\end{table}

As shown in Algorithm \ref{alg:rl_network_training}, the actor and critic networks are trained in multiple epochs. First, the actor and critic networks are randomly initialized (Line 1), and the initial bit-width allocation strategy $\mathcal{Q}_0$ is initialized to the maximum value for each layer (Line 2). Within each training epoch, the bit-width $q_t^t$ is generated for each layer (Lines 4-5).  The bit-width allocation strategy $\mathcal{Q}_t$ is updated with $q_t^t$ (Line 6). Then, the loss function corresponding to $\mathcal{Q}_t$ is computed (Line 7). Afterwards, the Temporal-Difference (TD) advantage is calculated according to Formula~\ref{eqn:advantage} (Line 8). In addition, the clip reward and the policy-dependent reward are calculated based on Formulas~\ref{eq:policy_reward} and~\ref{eq:clip_reward} (Lines 9-10). Finally, the critic network $\phi_t$ and the actor network $\theta_t$ are updated based on Formulas~\ref{eq:upgrade_theta} and~\ref{eqn:upgrade_phi} (Lines 11-12).

\subsection{Experiment Details}

In this section, we first present the hyperparameter values in experimental setup. Then, we present additional experiments, including the PPL with 5-bit quantization, varying average quantization bit-widths (from 3.1 to 4.5), the comparison of time consumption, and the quantization with 5-bits on average for LLaMA2-7B-chat, LLaMa2-13B-chat, and Mistral-7B-v0.1.

\subsubsection{Calibration-Size Robustness}
\label{subsubsec:calibration_size_robustness}

We examine whether the Fisher estimate and the resulting bit-width allocation are sensitive to the amount of calibration data. On LLaMA-7B, we construct nested C4 subsets containing 32, 64, 128, and 256 sequences for each of three independent seeds, and use the 256-sequence subset from the same seed as the reference. The analysis covers all 224 quantizable linear modules. To isolate calibration noise, every setting uses the same bit-conditioned proxy, candidate set $\{2,3,4\}$, parameter-weighted 3-bit budget, and deterministic same-budget allocator. Table~\ref{tab:calibration_size_robustness} reports the rank correlation, overlap among the top 10\% most sensitive modules, and the fraction of module assignments that differ from the 256-sequence reference.

\begin{table}[t]
    \centering
    \vspace{-2mm}
    \caption{Robustness to calibration-set size on LLaMA-7B over three seeds. Each row is compared with the nested 256-sequence subset from the same seed.}
    \label{tab:calibration_size_robustness}
    \vspace{-2mm}
    \renewcommand{\arraystretch}{1.08}
    \resizebox{\hsize}{!}{
    \begin{tabular}{cccc}
        \toprule
        \textbf{C4 sequences} & \textbf{Spearman $\uparrow$} & \textbf{Top-10\% overlap $\uparrow$} & \textbf{Allocation diff. $\downarrow$} \\
        \midrule
        32  & $0.996 \pm 0.001$ & $97.1\% \pm 2.5\%$  & $2.68\% \pm 0.45\%$ \\
        64  & $0.997 \pm 0.001$ & $98.6\% \pm 2.5\%$  & $2.08\% \pm 0.68\%$ \\
        128 & $0.998 \pm 0.001$ & $98.6\% \pm 2.5\%$  & $1.34\% \pm 1.34\%$ \\
        256 & $1.000 \pm 0.000$ & $100.0\% \pm 0.0\%$ & $0.00\% \pm 0.00\%$ \\
        \bottomrule
    \end{tabular}
    }
\end{table}

Even with only 32 calibration sequences, the Fisher ranking retains a Spearman correlation of 0.996 and a 97.1\% top-10\% overlap with the 256-sequence reference, while only 2.68\% of module assignments change. At the 128-sequence setting used in the main experiments, the allocation difference decreases to 1.34\%. These results show that the sensitivity ranking and budget-constrained allocation are stable with limited calibration data. This study uses 512-token calibration sequences and a deterministic proxy allocator, and does not rerun final-backend PPL for every calibration size; therefore, it establishes ranking and allocation stability rather than complete invariance of downstream quality.

\subsubsection{PPL with Diverse Quantization Bit-width}

As shown in Table \ref{tab:Different_strategy}, the PPL corresponding to the quantization of 4-bit with one layer quantized to 5 bits can vary across different layers. MLP layers, especially the down-projection (mlp.down), correspond to significant PPL drop (up to 0.115 PPL drop). Among attention layers, the value projection (attn.v) corresponds to the highest sensitivity (up to 0.047). \name{} aligns with this diversity and thus yields strong performance.

\subsubsection{Diverse Average Quantization Bit-width}

\name{} can achieve varying average quantization bit-widths through adaptive layer-wise bit-width allocation strategies. As shown in Table \ref{tab:Adaptive Storage}, \name{} can achieve average bit-widths from 3.1 to 4.5, with correspondingly decreasing PPL.

\subsubsection{Comparison of Time Consumption}

As shown in Table \ref{tab:time consumption}, the quantization time of \name{} is comparable to several baselines and can be shorter than GPTQ (up to 37\%), GPTQv2 (up to 61\%), OWQ (up to 35\%), and OmniQuant (up to 66\%). Although RTN and AWQ can be faster than \name{} by up to 97\% and 46\%, respectively, they may incur substantially larger performance degradation under aggressive compression.

\begin{table*}[t]
    \centering
    \vspace{-4mm}
    \caption{PPL $\downarrow$ comparison on Qwen2.5, Qwen2.5-14B, and Mistral with 4-bit average quantization. ``Avg PPL'' denotes the average perplexity over Wiki2, PTB, and C4.}
    \label{tab:ppl_qwen_mistral_4bit}
    \vspace{-4mm}
    \renewcommand{\arraystretch}{1}
    \resizebox{\hsize}{!}{
    \begin{tabular}{l|c|cccc|cccc|cccc}
        \toprule
        \textbf{Model} &  &
        \multicolumn{4}{c|}{\textbf{Qwen2.5-7B}} &
        \multicolumn{4}{c|}{\textbf{Qwen2.5-14B}} &
        \multicolumn{4}{c}{\textbf{Mistral-7B-v0.1}} \\
        \midrule
        \textbf{Method} & \multicolumn{1}{c|}{Avg bit} &
        Wiki2 & PTB & C4 & Avg PPL &
        Wiki2 & PTB & C4 & Avg PPL &
        Wiki2 & PTB & C4 & Avg PPL \\
        \midrule
        FP16 & 16 &
        6.84 & 12.79 & 11.88 & 10.50 &
        5.29 & 10.87 & 10.35 & 8.84 &
        5.25 & 9.94 & 8.38 & 7.86 \\
        \midrule
        RTN & 4 &
        9.14 & 16.57 & 15.29 & 13.67 &
        6.85 & 12.85 & 11.98 & 10.56 &
        6.00 & 11.47 & 9.47 & 8.98 \\
        GPTQ & 4 &
        7.29 & 13.41 & 12.50 & 11.07 &
        5.84 & 11.36 & 10.81 & 9.34 &
        5.45 & 10.38 & 8.65 & 8.16 \\
        GPTQv2 & 4 &
        7.20 & 13.24 & \underline{12.21} & 10.88 &
        5.82 & 11.19 & 10.63 & 9.21 &
        5.43 & 10.25 & 8.60 & 8.09 \\
        OmniQuant & 4 &
        7.12 & 13.24 & \underline{12.21} & 10.86 &
        5.72 & 11.18 & \underline{10.62} & \underline{9.17} &
        / & / & / & / \\
        OWQ & 4 &
        7.26 & 13.32 & 12.43 & 11.00 &
        5.78 & 11.20 & 10.66 & 9.21 &
        5.44 & 10.28 & 8.62 & 8.11 \\
        AWQ & 4 &
        \underline{7.09} & \underline{13.22} & 12.23 & \underline{10.85} &
        \textbf{5.70} & \underline{11.17} & 10.63 & \underline{9.17} &
        \underline{5.39} & \textbf{10.19} & \underline{8.57} & \underline{8.05} \\
        \name{} & 4 &
        \textbf{7.07} & \textbf{13.17} & \textbf{12.20} & \textbf{10.81} &
        \textbf{5.70} & \textbf{11.15} & \textbf{10.61} & \textbf{9.15} &
        \textbf{5.37} & \textbf{10.19} & \textbf{8.55} & \textbf{8.04} \\
        \bottomrule
    \end{tabular}
    }
\end{table*}

\begin{table*}[t]
    \centering
    \vspace{-4mm}
    \caption{{The accuracy $\uparrow$ of quantized Qwen2.5-7B and Qwen2.5-13B on zero-shot reasoning tasks. \textbf{Bold} indicates the highest accuracy and \underline{underlined} indicates the second-highest. ``Avg bit'' represents the average width-bit. ``Avg acc'' represents the average accuracy of the 5 tasks.}}
    \label{tab:Zero-shot_accuracy_qwen}
    \vspace{-4mm}
    \renewcommand{\arraystretch}{1.2}
    \resizebox{\hsize}{!}{
    \begin{tabular}{l|c|cccccc|cccccc}
        \toprule
        \textbf{Model} & \multicolumn{1}{c}{}& \multicolumn{6}{c|}{\textbf{Qwen2.5-7B}} &
        \multicolumn{6}{c}{\textbf{Qwen2.5-14B}} \\
        \midrule
        \textbf{Method} & \multicolumn{1}{c|}{Avg bit} & BoolQ & ARC-E & ARC-C & HellaSwag & WinoGrande & Avg acc  & BoolQ & ARC-E & ARC-C & HellaSwag & WinoGrande & Avg acc  \\
        \midrule
        FP16 & 16 & 0.8471 & 0.8047 & 0.4778 & 0.6003 & 0.7301 & 0.6920 &0.8522 &0.8244 &0.5597 &0.6338 &0.7537 &0.7248 \\
        \midrule
        RTN & 4 & 0.7883 & 0.7415 & 0.4377 & 0.5554 & 0.6645 & 0.63748 & 0.8144&	0.7988&	0.5042&	0.6088&	0.6921&	0.6836 \\
        GPTQ & 4 & 0.8394 & \underline{0.7988} & \underline{0.4692} & 0.5913 & 0.7111 & 0.68196 & 0.8404&	\underline{0.8232}&	\underline{0.5546}&	0.6241&	0.7334&	0.7151 \\
        GPTQv2 & 4 & \underline{0.8421}&0.7974&	0.4661&	\underline{0.5923}&	\underline{0.7139}&	\underline{0.68236}& \underline{0.8469}&	0.8167	&0.5527	&0.6244	&0.7329	&0.7147\\
        OmniQuant & 4 &0.8132&	0.7881&	0.4679&	0.5912&	0.7104&	0.67416& 0.8454&	0.8223	&0.5475&	0.6263	& \underline{0.7568}&	0.7197\\
        OWQ & 4 & 0.8012&	0.7832&	0.4521&	0.5723&	0.6985&	0.66146 & 0.8435&	0.8123&	0.5316&	0.6183& 0.7268&	0.7065\\
        AWQ & 4 & 0.8143& 0.7958&	0.4650&	\textbf{0.5926}&	\textbf{0.7150}&	0.67654 & 0.8391&	\textbf{0.8274}	& \textbf{0.5614}	& \underline{0.6267}	&0.7537	&\underline{0.7217} \\
        \name{} (Ours) & 4 & \textbf{0.8495}&	\textbf{0.7996}&	\textbf{0.4812}&	0.5853&	0.6992&	\textbf{0.68296} & \textbf{0.8496}&	\textbf{0.8274}&	0.5511&	\textbf{0.6274}&	\textbf{0.7576}&	\textbf{0.7226} \\
        \bottomrule
    \end{tabular}
    }
\end{table*}

As shown in Table~\ref{tab:preprocessing_time}, the preprocessing stage of \name{}, which consists of per-layer Fisher sensitivity computation and an RL-based bit-width optimization search, is conducted on NVIDIA RTX 4090 GPUs. 
The sensitivity computation accounts for the majority of the preprocessing cost and scales with model size, whereas the RL-based search is lightweight, requiring only tens of seconds on a single GPU. Overall, the total preprocessing time remains below 70 minutes even for 14B-scale models, demonstrating the practical efficiency and scalability of \name{}.

\subsubsection{Results of LLaMA2, Qwen2.5 and Mistral-7B-v0.1 models}
\label{subsubsec:results}

\begin{table}[t]
    \centering
    \vspace{-4mm}
    \caption{{Comparison of perplexity results $\downarrow$ of different 4-bit quantization approaches with the LLaMA2-7B-chat and LLaMA2-13B-chat model. ``Avg PPL'' represents the average PPL of the 3 benchmarks. }}
    \label{tab:perplexity of 4bit llama2}
    \vspace{-4mm}
    \renewcommand{\arraystretch}{1}
    \resizebox{\hsize}{!}{
    \begin{tabular}{l|c|cccc|cccc}
        \toprule
        \textbf{Model} & \multicolumn{1}{c}{}& \multicolumn{4}{c|}{\textbf{LLaMA2-7B-chat}} &
        \multicolumn{4}{c}{\textbf{LLaMA2-13B-chat}} \\
        \midrule
        \textbf{Method} & \multicolumn{1}{c|}{Avg bit} & Wiki2 & PTB & C4 & Avg PPL& Wiki2 & PTB & C4 & Avg PPL\\
        \midrule
        FP16 & 16 & 6.94 & 12.07 & 9.51 & 9.51& 5.09 & 9.08 & 6.79 & 6.99\\
        \midrule
        RTN & 4 & 7.96 & 13.70 & 10.94 & 10.87 & 6.42 & 11.05 & 9.02 & 8.83\\
        GPTQ & 4 & 7.29 & 12.76 & 10.12& 10.06 & 6.29  & 10.93 & 8.78 & 8.67\\
        GPTQv2 & 4 & 7.16 & 12.51 & \textbf{9.75} & \underline{9.81}& \underline{6.23} & 10.90 & 8.69 & 8.61\\ 
        OmniQuant & 4 & 7.15 & \underline{12.45} & 9.91 & 9.84& 6.27 & 10.99 & 8.82 & 8.69\\
        OWQ &  4 & 7.22 & 12.56 & 10.03 &9.94 & 6.25 & 10.89 & \underline{8.64} & 8.59\\
        AWQ &  4 & \underline{7.15} & 12.48 & 9.85 &9.83& \textbf{6.21} & \underline{10.87} & 8.65 & \underline{8.58}\\
        \name{} (Ours) & 4 & \textbf{7.12} & \textbf{12.41} & \underline{9.82} &\textbf{9.78}& \textbf{6.21} & \textbf{10.85} & \textbf{8.63} & \textbf{8.56} \\
        \bottomrule
    \end{tabular}
    }
\end{table}

As shown in Table \ref{tab:perplexity of 4bit llama2}, \name{} achieves the lowest average PPL under 4-bit quantization for both LLaMA2-7B-chat and LLaMA2-13B-chat (up to 1.09 lower than RTN, 0.28 lower than GPTQ, 0.05 lower than GPTQv2, 0.13 lower than OmniQuant, 0.16 lower than OWQ, and 0.05 lower than AWQ). The average reduction over all quantized baselines is larger on LLaMA2-7B-chat (0.28 PPL) than on LLaMA2-13B-chat (0.10 PPL), while both model sizes show the best average PPL with \name{}. Furthermore, \name{} outperforms baseline approaches for the majority of the combinations of the benchmarks and models (up to 0.84 for Wiki2 and 1.29 for PTB in LLaMA2-7B-chat; up to 0.21 for Wiki2, 0.20 for PTB, and 0.39 for C4 in LLaMA2-13B-chat). While the PPL of \name{} is slightly (0.07) higher than that of GPTQv2 with the combination of C4 and LLaMA2-7B-chat, \name{} outperforms other baselines in this setting (1.12 lower than RTN, 0.30 lower than GPTQ, 0.09 lower than OmniQuant, 0.21 lower than OWQ, and 0.03 lower than AWQ).

\begin{figure}[!t]
\centering
\includegraphics[width=0.85\linewidth]{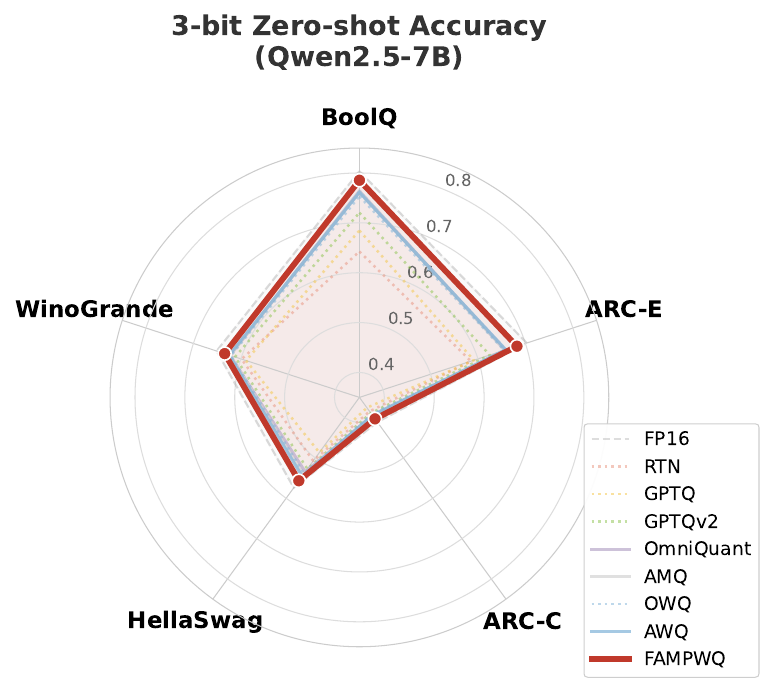}
\captionsetup{width=\linewidth}
\vspace{-4mm}
\caption{Zero-shot accuracy comparison under 3-bit quantization on Qwen2.5-7B. \name{} (red) consistently outperforms baselines across five reasoning tasks.}
\label{fig:radar_3bit}
\vspace{-8mm}
\end{figure}

As shown in Table \ref{tab:ppl_qwen_mistral_4bit}, \name{} consistently achieves the best performance in terms of PPL across all 3 LLMs and 3 benchmarks under 4-bit quantization. With Qwen2.5-7B, \name{} attains an average PPL of 10.81, which is up to 2.86 lower than that of RTN, and also lower than GPTQ, GPTQv2, OmniQuant, OWQ, and AWQ by 0.26, 0.07, 0.05, 0.19, and 0.04, respectively. On Qwen2.5-14B, \name{} achieves an average PPL of 9.15, which is 1.41 lower than RTN, 0.19 lower than GPTQ, 0.06 lower than GPTQv2, and 0.02 lower than both OmniQuant and AWQ. For Mistral-7B-v0.1, \name{} yields an average PPL of 8.04, representing reductions of 0.94, 0.12, 0.05, 0.07, and 0.01 compared to RTN, GPTQ, GPTQv2, OWQ, and AWQ, respectively.

As shown in Table \ref{tab:Zero-shot_accuracy_qwen}, \name{} significantly outperforms baseline approaches (from 0.06\% to 4.55\%) in terms of average accuracy with Qwen2.5-7B and Qwen2.5-14B under 4-bit quantization. As shown in Table \ref{tab:Zero-shot_accuracy_llama}, \name{} achieves the highest average zero-shot accuracy under 4-bit quantization with both LLaMA2-7B-chat (up to 1.63\% higher than RTN, 0.39\% higher than GPTQ, and 0.2\% higher than AWQ) and LLaMA2-13B-chat (up to 2.87\% higher than RTN), with \name{} surpassing all or most baseline approaches for the majority of tasks.

\begin{table*}[t]
    \centering
    \vspace{-4mm}
    \caption{{The accuracy $\uparrow$ of quantized LLaMA2-7B-chat and LLaMA2-13B-chat on 5 zero-shot reasoning tasks under 4-bit quantization. \textbf{Bold} indicates the highest accuracy and \underline{underlined} indicates the second-highest. ``Avg bit'' represents the average width-bit. ``Avg acc'' represents the average accuracy of the 5 tasks.}}
    \label{tab:Zero-shot_accuracy_llama}
    \vspace{-4mm}
    \renewcommand{\arraystretch}{1.2}
    \resizebox{\hsize}{!}{
    \begin{tabular}{l|c|cccccc|cccccc}
        \toprule
        \textbf{Model} & \multicolumn{1}{c}{}& \multicolumn{6}{c|}{\textbf{LLaMA2-7B-chat}} &
        \multicolumn{6}{c}{\textbf{LLaMA2-13B-chat}} \\
        \midrule
        \textbf{Method} & \multicolumn{1}{c|}{Avg bit} & BoolQ & ARC-E & ARC-C & HellaSwag & WinoGrande & Avg acc & BoolQ & ARC-E & ARC-C & HellaSwag & WinoGrande & Avg acc \\
        \midrule
        FP16 & 16 & 0.8034 & 0.7028 & 0.4112 & 0.5740 & 0.6511 & 0.6285 &0.8302	&0.7571	&0.447	&0.6061	&0.7103	&0.6701 \\
        \midrule
        RTN & 4 & 0.7425 & 0.6851 & 0.4018 & 0.5521 & 0.6551 & 0.6073 & 0.8125	&0.7313	&0.4085	&0.5577	&0.6787	&0.6377\\
        GPTQ & 4 & \underline{0.8017} & 0.6866 & 0.4052 & 0.5627 & 0.6421 & 0.6197
        & \underline{0.8220}& 0.7521& \underline{0.4436}&0.5926 &	\underline{0.7127}&	{0.6646} \\
        GPTQv2 & 4 & 0.8015	&0.6883	&0.4064	&0.5636	&\textbf{0.6477}	&0.6215 & 0.8201&	\underline{0.7530}&	0.4432&	\underline{0.5993}&	0.7083&	\underline{0.6648}\\
        OmniQuant & 4 &\textbf{0.8021}&\textbf{0.6975}	&0.4071	&\underline{0.5688}	&0.6418 &\underline{0.6234}& 0.8204	&0.7482	&0.4324	&0.5946	&0.7034	&0.6598\\
        OWQ & 4 & 0.7953 & 0.6898 & \underline{0.4083} & 0.5642 & 0.6422 & 0.6199 & 0.8213&	\textbf{0.7542}&0.4410&	0.5892&	0.7078&	0.6627 \\
        AMQ & 4 & 0.7963 & 0.6894 & 0.4067 & 0.5655 & 0.6423 & 0.6200 & 0.8208 & 0.7523 & 0.4392 & 0.5964 & 0.7008 & 0.6619 \\
        AWQ & 4 & 0.7975 & 0.6948 & 0.4069 & 0.5672 & 0.6416 & 0.6216 & 0.8217&	0.7478&	0.4431&	0.5937&	\textbf{0.7166}&	0.6645\\
        \name{} (Ours) & 4 & 0.8012 & \underline{0.6957} & \textbf{0.4095} & \textbf{0.5692} & \underline{0.6424} & \textbf{0.6236} & \textbf{0.8244}	& 0.7474&	\textbf{0.4453}&	\textbf{0.6023}&	0.7127&	\textbf{0.6664} \\
        \bottomrule
    \end{tabular}
    }
\end{table*}

\subsubsection{Storage reduction}
Comparable to single-precision quantization approaches, e.g., RTN, AWQ, GPTQ, GPTQv2, \name{} incurs no extra storage overhead or no additional metadata, while OWQ corresponds to larger storage requirement with extra metadata. As shown in the Table \ref{tab:model_storage_reduction}, \name{} reduces storage overhead by 1\%–3\% of the original model size compared to OWQ. As the quantization bit width of the model weights decreases, the storage space required by each model is almost linearly reduced.

\begin{table}[t]
    \centering
    \vspace{-4mm}
    \caption{Storage requirements of various LLMs under different weight quantization precisions. Percentages in parentheses indicate the proportion relative to FP16 size.}
    \label{tab:model_storage_reduction}
    \vspace{-4mm}
    \renewcommand{\arraystretch}{1}
    \resizebox{\hsize}{!}{
    \begin{tabular}{l|c|c|c|c|c|c|c}
        \toprule
        \multirow{2}{*}{\textbf{Model}} & &\multicolumn{3}{c|}{\textbf{RTN/AWQ/GPTQ\&v2/\name{} Storage (MB)}} &\multicolumn{3}{c}{\textbf{OWQ Storage (MB)}} \\
        \cmidrule(lr){2-8}
        & \textbf{FP16} & \textbf{5-bit} & \textbf{4-bit} & \textbf{3-bit} & \textbf{5-bit} & \textbf{4-bit} & \textbf{3-bit}\\
        \midrule
        LLaMA-7B            & 12,853 & 4,420 (34\%) & 3,589 (28\%) & 2,817 (22\%) 
        & 4,505 (35\%) & 3,674 (29\%) & 2,899 (23\%)\\
        LLaMA-13B           & 24,826 & 8,550 (34\%) & 6,676 (27\%) & 5,164 (21\%) 
        & 8,635 (35\%) & 6,761 (27\%) & 5,249 (21\%) \\
        LLaMA2-7B-chat      & 12,853 & 4,420 (34\%) & 3,589 (28\%) & 2,817 (22\%) 
        & 4,505 (35\%) & 3,674 (29\%) & 2,899 (23\%)\\
        LLaMA2-13B-chat     & 24,826 & 8,550 (34\%) & 6,676 (27\%) & 5,164 (21\%) 
        & 8,635 (35\%) & 6,761 (27\%) & 5,249 (21\%)\\
        Qwen2.5-7B          & 15,317 & 5,800 (38\%) & 5,197 (34\%) & 4,414 (29\%) 
        & 5,883 (38\%) & 5,280 (34\%) & 4,502 (29\%) \\
        Qwen2.5-14B         & 28,172 & 10,600 (38\%) & 9,272 (33\%) & 7,697 (27\%) 
        &10,680 (38\%) & 9,365 (33\%) & 7,782 (28\%) \\
        Mistral-7B-v0.1     & 13,825 & 4,750 (34\%) & 3,841 (28\%) & 3,009 (22\%) 
        & 4,826 (35\%) & 3,920 (28\%) & 3,075 (22\%)\\
        \bottomrule
    \end{tabular}
    }
\end{table}

\subsubsection{Sensitivity Metric Comparison}
The sensitivity metric comparison is visualized in the main text (Figure~\ref{fig:sensitivity_metric}). Table~\ref{tab:sensitivity_comparison} provides the same data in an extended format for reference.

\begin{table}[ht]
    \centering
    \vspace{-4mm}
    \caption{Sensitivity analysis of different metrics on LLaMA-7B and Qwen2.5-7B (extended from the main-text sensitivity comparison).}
    \label{tab:sensitivity_comparison}
    \vspace{-4mm}
    \renewcommand{\arraystretch}{1.2}
    \setlength{\tabcolsep}{6pt}
    \resizebox{\hsize}{!}{
    \begin{tabular}{lcccccc}
        \toprule
        \multirow{2}{*}{\textbf{Metric}} & \multicolumn{3}{c}{LLaMA-7B} & \multicolumn{3}{c}{Qwen2.5-7B} \\
        \cmidrule(lr){2-4} \cmidrule(lr){5-7}
         & $r \uparrow$ & Final PPL $\downarrow$ & $\Delta$PPL & $r \uparrow$ & Final PPL $\downarrow$ & $\Delta$PPL \\
        \midrule
        Random Allocation & 0.04 & 6.87 & +1.19 & 0.02 & 8.31 & +1.47 \\
        Weight Magnitude ($\|W\|_2$) & 0.42 & 6.53 & +0.85 & 0.38 & 8.06 & +1.22 \\
        \textbf{FIM} & \textbf{0.91} & \textbf{6.10} & \textbf{+0.42} & \textbf{0.88} & \textbf{7.63} & \textbf{+0.79} \\
        \midrule
        Oracle (Ground-truth) & 1.00 & 6.02 & +0.34 & 1.00 & 7.51 & +0.67 \\
        \bottomrule
    \end{tabular}
    }
\end{table}

\subsubsection{Packed-Deployment Memory Accounting}
\label{subsubsec:packed_memory_accounting}

Static model size alone does not capture the complete deployment footprint. We therefore perform analytical tensor accounting for packed Llama-2-7B inference with a 512-token prompt, 256 generated tokens, and an FP16 KV cache. The packed static weight footprints are 12.551~GiB for FP16, 2.862~GiB at a 3-bit average, and 3.622~GiB at a 4-bit average. Because \name{} changes only the weight representation, its KV-cache footprint is identical to AWQ: 0.375~GiB at batch size 1 and 1.500~GiB at batch size 4. The largest per-layer FP16 materialization is at most 86~MiB (0.084~GiB), and this workspace is reused across sequential layer execution rather than allocated once per layer.

\begin{table}[t]
    \centering
    \vspace{-2mm}
    \caption{Analytical packed-deployment memory accounting for Llama-2-7B. Values include packed weights, the final FP16 KV cache, and the known reusable workspace; they are not measured runtime peaks.}
    \label{tab:packed_memory_accounting}
    \vspace{-2mm}
    \renewcommand{\arraystretch}{1.08}
    \resizebox{\hsize}{!}{
    \begin{tabular}{cccccc}
        \toprule
        \textbf{Avg. bit} & \textbf{Batch} & \textbf{FP16} & \textbf{AWQ} & \textbf{\name{} upper bound} & \textbf{Saving vs. FP16} \\
        \midrule
        3 & 1 & 12.926~GiB & 3.237~GiB & $\leq 3.321$~GiB & $\geq 9.605$~GiB \\
        3 & 4 & 14.051~GiB & 4.362~GiB & $\leq 4.446$~GiB & $\geq 9.605$~GiB \\
        4 & 1 & 12.926~GiB & 3.997~GiB & $\leq 4.081$~GiB & $\geq 8.845$~GiB \\
        4 & 4 & 14.051~GiB & 5.122~GiB & $\leq 5.206$~GiB & $\geq 8.845$~GiB \\
        \bottomrule
    \end{tabular}
    }
\end{table}

At the 3-bit average budget, packing reduces the weight footprint by 9.689~GiB, whereas the largest known temporary mixed-bit workspace is only 0.084~GiB. The resulting tensor-accounted saving is therefore at least 9.605~GiB, and the workspace is only 0.87\% of the static weight saving. Under this accounting, temporary mixed-bit storage cannot offset the weight-memory reduction. This result is an analytical estimate rather than a measured runtime peak: activation and framework residuals, CUDA-reserved memory, allocator behavior, and fragmentation still require measurement with an actual packed-kernel implementation.

\subsubsection{Bit-width Allocation Visualization}

\begin{figure*}[!t]
\centering
\includegraphics[width=\linewidth]{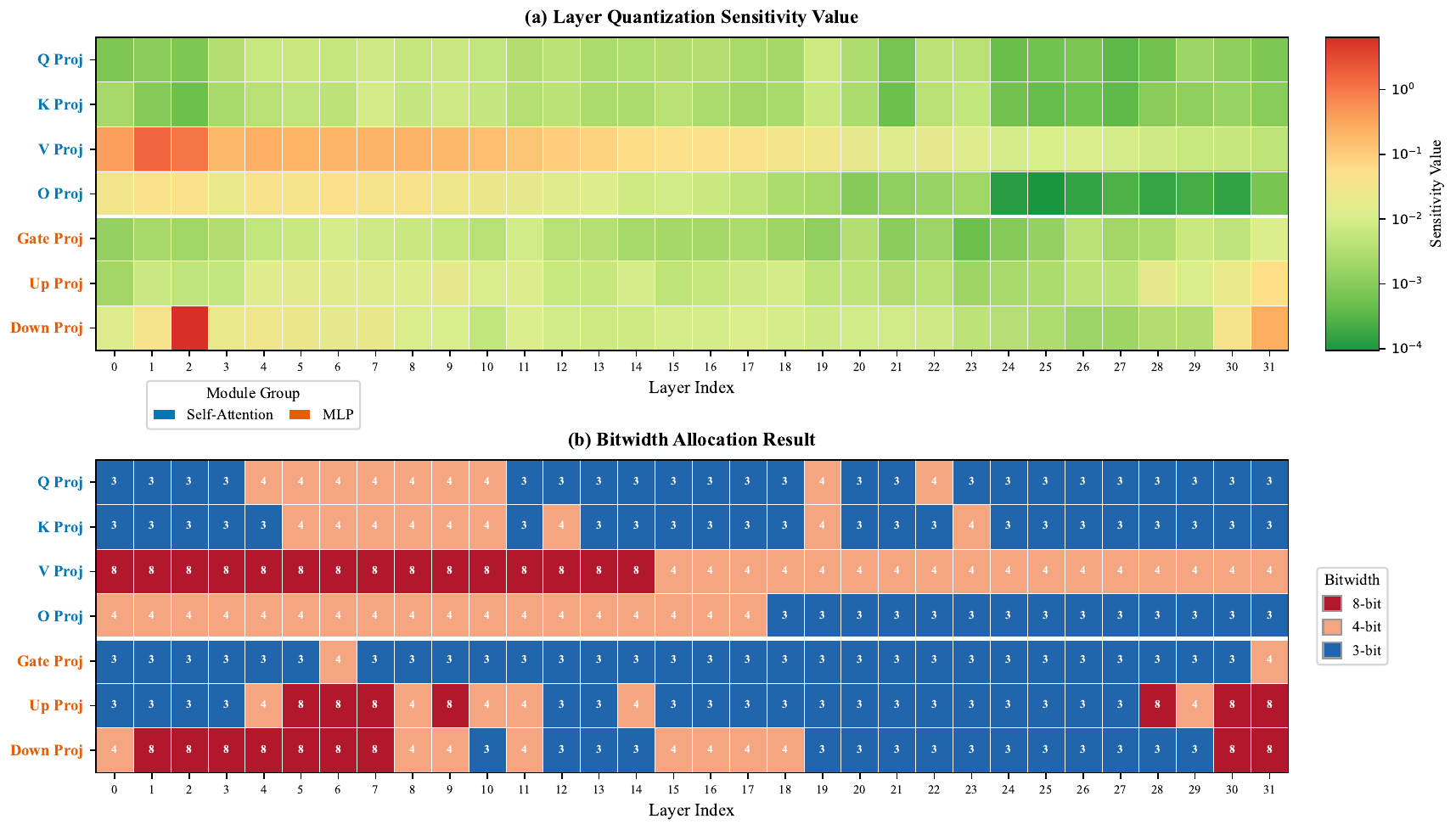}
\captionsetup{width=\linewidth}
\vspace{-4mm}
\caption{Layer-wise sensitivity scores (top) and the resulting bit-width allocations (bottom) under the three metrics on LLaMA-7B ($b_t=4$). }
\label{fig:sensitivity_and_allocation}
\vspace{-4mm}
\end{figure*}

Fig.~\ref{fig:sensitivity_and_allocation} visualizes the per-layer sensitivity scores and the corresponding quantization-optimized bit-width allocations. FIM captures gradient-level information that reveals additional critical layers overlooked by weight-only metrics. Thus, \name{} can generate a structurally distinct allocation that assigns higher precision to the most loss-sensitive modules. In this way, \name{} ultimately delivers excellent performance.

\subsubsection{Validation of Modeling Assumptions}
\label{subsubsec:modeling_assumptions}

Our accuracy-degradation proxy (Formulas~\ref{eq:accuracy_loss2single_layer}--\ref{eq:accuracy_loss2total_model}) rests on two assumptions: (1) exponential decay of degradation with bit-width, and (2) approximate layer independence.

\textbf{Exponential Decay.} Following Rate-Distortion Theory \cite{zhou2018adaptive}, quantization error decreases exponentially with allocated bits. We verify this empirically by measuring per-layer PPL as a function of bit-width and fitting exponential curves ($R^2 > 0.95$ across all layer types for LLaMA-7B; see Figure~\ref{fig:alpha_sensitivity_curve}).

\textbf{Layer Independence.} To validate the additivity assumption, we perform a controlled test on LLaMA-7B: we quantize layer pairs $(L_{10}, L_{11})$ and $(L_5, L_{25})$ individually, sum their PPL increases, and compare against the joint quantization. The relative error between the additive prediction and actual degradation is $<$0.1\%, confirming that cross-layer interaction effects are negligible for the purpose of bit-width allocation.

\textbf{$\alpha$ Sensitivity.} The decay rate $\alpha$ in Formula~\ref{eq:accuracy_loss2single_layer} is robust across a wide range: as shown in Figure~\ref{fig:alpha_sensitivity_curve}, PPL varies by less than 0.3 within $\alpha \in [15, 25]$ for both LLaMA-7B and Qwen2.5-7B.

\paragraph{Controlled Error Additivity Test.}
\label{subsubsec:error_additivity}

To further validate the layer-independence assumption used in the proxy model, we quantize two layers individually and jointly while keeping all other layers at FP16. If cross-layer interactions are negligible, the joint PPL increase should match the sum of the two individual increases.

\begin{table}[t]
    \centering
    \vspace{-2mm}
    \caption{Controlled error additivity test on LLaMA-7B under 3-bit quantization. Relative error compares the additive prediction $\Delta(L_i)+\Delta(L_j)$ against the jointly measured degradation $\Delta(L_i,L_j)$.}
    \label{tab:error_additivity}
    
    \renewcommand{\arraystretch}{1.1}
    \resizebox{\hsize}{!}{
    \begin{tabular}{l|c|c|c|c|c}
        \toprule
        \textbf{Layers} & $\Delta(L_i)$ & $\Delta(L_j)$ & \textbf{Sum} & \textbf{Actual} & \textbf{Rel. error} \\
        \midrule
        $L_{10}, L_{11}$ (Adjacent) & 0.0084 & 0.0079 & 0.0163 & 0.016315 & 0.09\% \\
        $L_{5}, L_{25}$ (Distant) & 0.0062 & 0.0112 & 0.0174 & 0.017412 & 0.07\% \\
        \bottomrule
    \end{tabular}
    }
\end{table}

As shown in Table~\ref{tab:error_additivity}, the relative error is below 0.1\% for both adjacent and distant layer pairs, supporting the approximation that cross-layer interaction effects are small for the purpose of bit-width allocation.

\begin{figure}[!t]
\centering
\includegraphics[width=\linewidth]{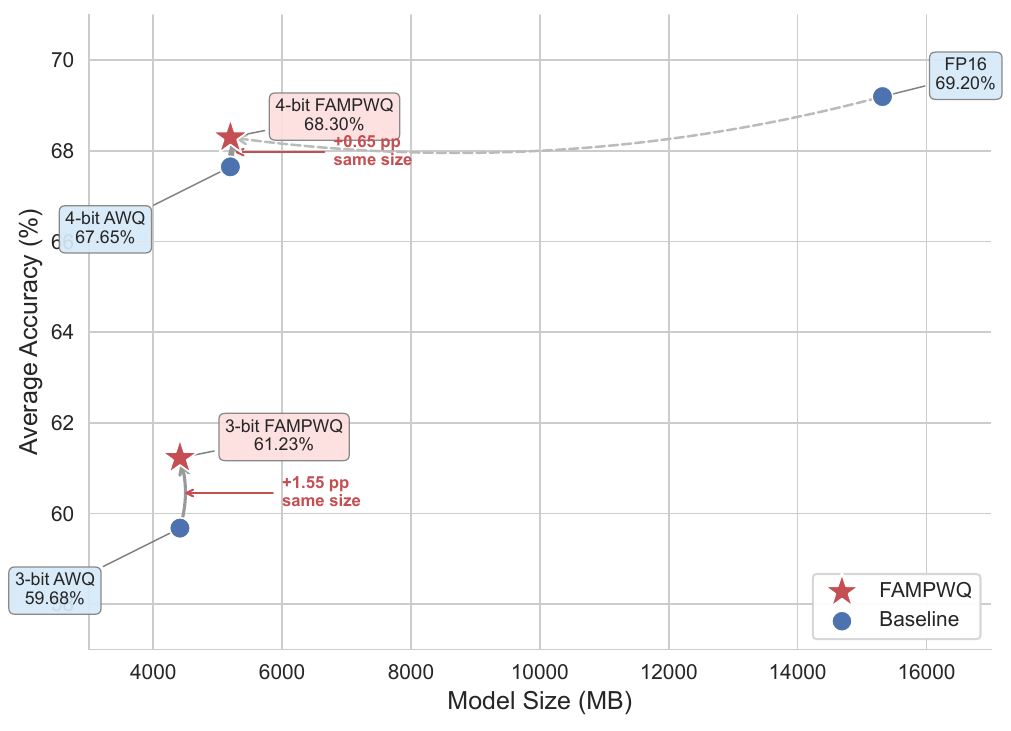}
\captionsetup{width=\linewidth}
\vspace{-4mm}
\caption{Accuracy vs.\ model size trade-off for Qwen2.5-7B. At identical storage cost, \name{} achieves +0.65 percentage points over AWQ at 4-bit and +1.55 points at 3-bit, demonstrating better accuracy-per-byte efficiency.}
\label{fig:accuracy_per_byte}
\vspace{-4mm}
\end{figure}

\subsubsection{Vicuna-Bench Generation Quality at 3-bit}
\label{subsubsec:vicuna_3bit}

To bridge perplexity and real-world generation quality, we also evaluate 3-bit quantized models on Vicuna-Bench with GPT-4 as the judge. Table~\ref{tab:vicuna_3bit} shows that \name{} preserves generation quality better than the strongest uniform baseline under this aggressive compression setting.

\begin{table}[t]
    \centering
    \vspace{-2mm}
    \caption{Vicuna-Bench generation quality under 3-bit quantization. Win rate is measured against FP16 responses.}
    \label{tab:vicuna_3bit}
    \vspace{-4mm}
    \renewcommand{\arraystretch}{1.1}
    \resizebox{0.75\hsize}{!}{
    \begin{tabular}{l|c|c}
        \toprule
        \textbf{Method} & \textbf{Win rate} & \textbf{Avg. length} \\
        \midrule
        RTN & 2.1\% & 164.2 \\
        GPTQ & 13.3\% & 193.8 \\
        AWQ & 18.5\% & 192.4 \\
        \textbf{\name{} (Ours)} & \textbf{32.3\%} & \textbf{208.0} \\
        \bottomrule
    \end{tabular}
    }
\end{table}

\subsubsection{Accuracy-per-Byte Analysis}

To evaluate the deployment efficiency of \name{}, we analyze the accuracy-per-byte trade-off. As shown in Figure~\ref{fig:accuracy_per_byte}, at identical storage costs (same average bit-width), \name{} consistently achieves higher accuracy than the best uniform baseline (AWQ). The advantage grows as compression becomes more aggressive: +0.65 percentage points at 4-bit (5,197 MB) and +1.55 points at 3-bit (4,414 MB). This confirms that \name{} extracts more quality from each byte of storage, making it particularly valuable for memory-constrained deployment scenarios where the goal is fitting the best possible model into a fixed VRAM budget.

\subsubsection{Qualitative Generation Example at 3-bit}
\label{subsubsec:qualitative_generation_example}

\begin{table}[t]
    \centering
    \caption{Representative Llama-2-Chat responses at a 3-bit average budget; light blue marks the factual error in 3-bit AWQ.}
    \label{tab:qualitative_generation_example}
    \vspace{-2mm}
    \scriptsize
    \setlength{\tabcolsep}{3pt}
    \renewcommand{\arraystretch}{1.05}
    \begin{tabular}{>{\centering\arraybackslash}m{0.16\hsize}p{0.74\hsize}}
        \toprule
        \textbf{Model} & \textbf{Generated response} \\
        \midrule
        FP16 & The cause of the delay is the introduction of a payment bug in the Tuesday update that can charge mobile orders twice. Seventeen beta users were affected, but desktop orders were not. The revised plan is to launch on Monday instead of Friday and to notify beta users and refund the affected users. The decision to launch on Monday will be made on Friday at 3 p.m. \\
        \addlinespace
        3-bit AWQ & The cause of the delay was a payment bug that affected 17 of the 120 beta users but not the desktop orders. The customer impact was that the bug caused the mobile orders to be charged twice. \textcolor{cyan!65!blue}{The team was able to patch the issue by Tuesday and the patch was completed by Thursday.} The revised plan was to move the launch to Monday and notify the affected beta users and refund them. The team will make the final go/no-go decision on Friday at 3 p.m. \\
        \addlinespace
        3-bit \name{} & The cause of the delay was a payment bug that charged mobile orders twice. This affected 17 beta users, but not the desktop users. The team was able to fix the issue by the end of the week, but the QA process will take two full days, so the launch will be on Monday. The decision to launch on Monday will be made on Friday at 3 p.m. \\
        \bottomrule
    \end{tabular}
\end{table}

Aggregate win rates do not reveal which information is lost by a quantized model. We therefore compare one representative response from FP16, 3-bit AWQ, and 3-bit \name{} on Llama-2-Chat with \texttt{max\_length=200}. The prompt requires a short summary of the cause, customer impact, and revised plan described in a multi-speaker dialogue:

\begin{quote}
\footnotesize
Briefly summarize the cause of the delay, the customer impact, and the revised plan described below. Write one short paragraph.

\textbf{Maya:} Can we still launch on Friday?\\
\textbf{Leo:} No. Tuesday's update introduced a payment bug that can charge mobile orders twice. Seventeen of our 120 beta users were affected; desktop orders were not affected. I can finish the patch by Thursday.\\
\textbf{Nina:} QA needs two full days after the patch, so we should move the launch to Monday. I will notify the beta users and refund the 17 affected users today.\\
\textbf{Maya:} Agreed. We will make the final go/no-go decision on Friday at 3 p.m.
\end{quote}

Light-blue text marks AWQ's factual timeline error: it places the patch on Tuesday, the day the bug was introduced, and omits the two-day QA period. \name{} preserves the cause, impact, QA delay, and Monday launch without introducing this contradiction, although it gives less precise patch timing and omits the notification and refund action. This example illustrates a specific low-bit failure mode rather than an aggregate claim.

\end{document}